\pdfoutput=1
\documentclass{article}
\usepackage{arxiv_style,times}
\usepackage[T1]{fontenc}
\usepackage{amsmath,amssymb,mathtools}
\usepackage{hyperref,url}
\usepackage{graphicx,wrapfig}
\usepackage{booktabs,multirow,colortbl,caption}
\usepackage{xcolor,microtype}
\usepackage{algorithm,algpseudocode}
\usepackage{tikz,pgfplots}
\pgfplotsset{compat=1.18}
\usepackage{dsfont,array,longtable}

\definecolor{bmteal}{HTML}{0E5C56}
\definecolor{bmviolet}{HTML}{8577AD}
\definecolor{bmochre}{HTML}{C98A2C}
\definecolor{bmterra}{HTML}{C1502C}
\definecolor{bmink}{HTML}{1A2226}
\definecolor{bminkgrey}{HTML}{4B5A61}
\definecolor{bmpanel}{HTML}{F4F6F7}
\definecolor{bmline}{HTML}{D8DEE0}
\definecolor{bmterrapanel}{HTML}{FBEFE9}

\hypersetup{
  colorlinks=true,
  linkcolor=bmteal,
  citecolor=bmteal,
  urlcolor=bmteal,
  linktoc=all,
  pdfborder={0 0 0}
}

\newcommand{\cD}{\mathcal{D}}
\newcommand{\cE}{\mathcal{E}}
\newcommand{\bbR}{\mathbb{R}}
\newcommand{\DP}{\mathrm{DP}}
\newcommand{\NIW}{\mathrm{NIW}}
\newcommand{\Cat}{\mathrm{Cat}}
\newcommand{\Nor}{\mathcal{N}}
\DeclareMathOperator*{\argmax}{arg\,max}
\DeclareMathOperator*{\argmin}{arg\,min}

\title{Test-Time Adaptation of Reasoning Strategies \\ with Bayesian Nonparametric Memory }

\author{Keshav Ramji\thanks{Correspondence to \texttt{keshav.ramji@ibm.com}}~, ~Tahira Naseem \& Ramón Fernandez Astudillo \\
IBM Research AI}

\begin{document}

\maketitle

\begin{abstract}

While modern large language models (LLMs) have been trained to reason through verbalized chains-of-thought, the generation cost grows substantially due to suboptimal paths to reach the final answer. Furthermore, as new insights are discovered while observing various input queries (e.g. through self-reflection), limited mechanisms exist for carrying forward these findings to be applied to subsequent problems. One can view the list of such strategies or behaviors as a growing cheatsheet, with elements retrieved from this memory module at inference-time. In this work, we consider \textit{structured} cheatsheets, with learned clusters of behaviors. We introduce a Hierarchical Dirichlet Process Gaussian Mixture Model (HDP-GMM) over behavior embeddings, which shares components across domains while allowing domain-specific mixing weights, and uses the posterior predictive to retrieve relevant behaviors for a query; we call this a \textit{Bayesian Cheatsheet}. This mechanism allows for cheap adaptation in an online test-time training (TTT) setting, softly updating the mixture's sufficient statistics following each sample and enabling the creation of new components when the synthesized behaviors are sufficiently novel. We demonstrate that Bayesian Cheatsheet achieves clear performance gains relative to existing memory modules across reasoning benchmarks such as AIME'25, Omni-MATH, and PhysReason, even in the cold-start setting. We show that the Bayesian Cheatsheet is an adaptively reorganizing memory module, as behaviors can be re-assigned to components through a single step of collapsed Gibbs sampling.
Our findings highlight the value of Bayesian-inspired memory modules for effective test-time adaptation and  the role of structure in metacognitive reasoning.
\end{abstract}

\par
\noindent\begin{minipage}{\linewidth}
\centering
\includegraphics[width=\linewidth]{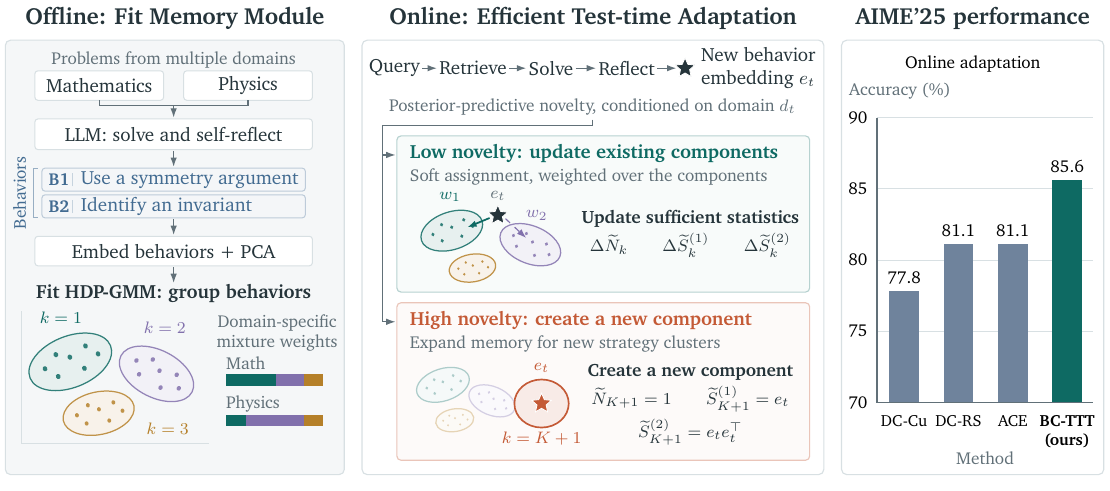}
\captionsetup{font=small,skip=4pt,hypcap=false}
\captionof{figure}{Bayesian Cheatsheet learns a structured memory module offline and adapts it at test time. Left: behaviors are mined offline by an LLM solving problems and self-reflecting on its solutions; the extracted behaviors are embedded and clustered into shared components with domain-specific weights by fitting an HDP-GMM. Middle: at test time, a novelty score based on the HDP-GMM's posterior predictive determines whether a new behavior updates existing components' sufficient statistics (top) or spawns a new component and initializes its sufficient statistics (bottom). Right: evaluation on AIME'25 with Granite-4.2-30B shows that Bayesian Cheatsheet (BC-TTT) outperforms the compared adaptive memory mechanisms in the online setting.}
\label{fig:memory-phases}
\end{minipage}
\par

\section{Introduction}

Large language models increasingly solve complex problems by generating verbalized intermediate reasoning steps \citep{wei2022cot, nye2021scratchpads}. Given that related problems can often be solved with similar strategies, storing such strategies, which we call \textit{behaviors}, can improve reasoning efficiency in solving future problems — that is, they do not need to be rediscovered. This is a key premise of metacognitive reasoning \citep{shinn2023reflexion,didolkar2024metacognitive, didolkar2025metacognitive}, where reflecting on reasoning to extract generalizable lessons and using those lessons to guide subsequent attempts have been shown to be effective. At the same time, this begs the question: how can we design systems to optimally retain and retrieve useful abstractions for LLM reasoning and decision-making?

Recent methods accumulate and revise reasoning behaviors, reflections, and procedural guidance over time \citep{didolkar2025metacognitive,suzgun2026cheatsheet, zhang2026ace, ouyang2026reasoningbank}. As memory grows, context limits and costs make it desirable to maintain an external, structured memory with selective reuse to avoid context explosion. Existing structured memory systems \citep{xu2025amem,chhikara2025mem0} often rely on LLM judgments and similarity-based heuristics to define and evolve the memory structure; this organization can become difficult to maintain as the memory grows.

In this work, we use the term \textit{cheatsheet} to refer to a general abstraction: it is a collection of concise, natural-language insights that can be consulted while solving a new problem. A cheatsheet induces two key design questions: 1) what type of information should be stored? and 2) how should it be organized for effective reuse and continual adaptation? In this work, we focus on the latter, while for the former, we simply use short reasoning behaviors generated through self-reflection to form the memory content. This allows for a clear comparison of memory organization mechanisms without the confounding effect of differences in the memory content. As a cheatsheet grows, its organization must support more than the retrieval of similar entries — related behaviors may form groups, which may also be reusable across domains. New experiences encountered at test-time may reinforce the contents of an existing group, or reveal a strategy that is not captured under the current organization. However, static cheatsheets, as well as flat similarity-based retrieval, do not address this, and while clustering provides additional organization for reuse, it does not facilitate principled growth and requires the number of clusters to be known in advance.

We introduce the Bayesian Cheatsheet, a structured memory module based on a Hierarchical Dirichlet Process Gaussian Mixture Model (HDP-GMM) \citep{rasmussen2000, teh2006} over behavior embeddings. The HDP-GMM formulation, as a non-parametric unbounded clustering mechanism, offers a few benefits that are particularly desirable for test-time memory management: 1) the number of clusters is purely driven by the data seen thus far, 2) the hierarchical structure offers a flexible mechanism to distinguish generally usable memories from domain-specific ones, and 3) the posterior probabilities of cluster assignment serve a dual purpose, in assigning memories to clusters as well as input-conditioned retrieval. The HDP prior additionally induces a rich-get-richer bias --
memory components repeatedly observed in experience become more likely to be used in the future, while new components may still be created.

Our Bayesian Cheatsheet memory learns globally shared components (clusters) of behaviors, allowing for probabilistically latent assignment. At inference time, it predicts the query's domain and retrieves behaviors from components to add to the language model's context.
In an online test-time training setting with streaming samples, Bayesian Cheatsheet cheaply incorporates newly mined behaviors into the memory through additive sufficient statistics, while also permitting the creation of new components for samples where the behaviors are sufficiently novel. Bayesian Cheatsheet also allows the memory to periodically reorganize itself as its contents evolve without requiring a refitting step every sample. As a result, this mechanism supports lightweight continual adaptation in a statistically-informed, training-free manner through principled changes to an external memory module.

\section{Related Work}\label{sec:related-work}

\paragraph{Test-time adaptation and continual learning.}
Early test-time training methods adapted models through self-supervised losses on test inputs \citep{sun2020ttt}, with later work on language models drawing adaptation data from retrieved text \citep{hardt2024tttllm} or input--output demonstrations supplied with a reasoning task \citep{akyurek2025ttt}. Continual learning additionally aims to retain previously learned capabilities across updates; Learning without Forgetting \citep{li2016lwf} and elastic weight consolidation \citep{kirkpatrick2017ewc} constrain changes in predictions and important parameters, respectively. On-policy distillation trains a student on its own generated sequences using a teacher's predictions \citep{agarwal2024gkd}; in self-distillation, the model itself supplies the teacher predictions when conditioned on richer information. Self-distillation fine-tuning uses demonstrations to acquire skills with less forgetting \citep{shenfeld2026}, and related methods distill guidance from textual feedback \citep{hubotter2026sdpo}, verified solutions \citep{zhao2026opsd}, or subsequent user messages \citep{kleinebuening2026interactions}. While these methods parametrically update language model weights, we focus on continual adaptation through an external memory.

\paragraph{Memory and retrieval-augmented reasoning with LLMs.}
Retrieval-augmented generation \citep{lewis2020rag} provides external evidence at inference time by retrieving relevant information from a corpus or memory store, with structures ranging from vector indexes to interconnected knowledge graphs \citep{gutierrez2025hipporag2}. As these memories grow, their contents must also be managed and revised: MemGPT manages information beyond the context window \citep{packer2024memgpt}, Mem0 consolidates conversational records \citep{chhikara2025mem0}, and A-MEM links and updates structured notes \citep{xu2025amem}. For reasoning, ExpeL and Metacognitive Reuse extract reusable guidance from prior problem solving \citep{zhao2024expel,didolkar2025metacognitive}, while Dynamic Cheatsheet, ACE, and ReasoningBank maintain and refine strategies as experience accumulates \citep{suzgun2026cheatsheet,zhang2026ace,ouyang2026reasoningbank}. Recent approaches also learn how memory is constructed or used: ALMA searches over executable memory designs \citep{xiong2026alma}, and MILES learns selection over an expanding library of step-level instructions \citep{tong2026miles}. The focus in this work is on the probabilistic organization of such memory content.

\section{Structured Retrieval for Metacognitive Reasoning}\label{sec:structured-retrieval}

\paragraph{Problem Setup.} We consider $J$ domains $\cD = \{1,\dots,J\}$ (in our experiments, math and physics), each with
a corpus of prompts. From these prompts we mine \emph{behaviors} -- short natural-language
reasoning strategies, extracted by having the model solve a problem, reflect on which general
strategies its solution used, and record them (details in Appendix~\ref{app:impl}) -- and embed
their descriptions with a fixed sentence encoder  $\cE:\mathcal{S}\to\bbR^{D_0}$ (where $\mathcal{S}$ is the space of text strings, including queries and behaviors), then project into a
lower-dimensional space via $f: \bbR^{D_0} \to \bbR^D$ (PCA, fit once over the mined
corpus). We use $e_{d,i} \in \bbR^D$ to denote the projected embedding of the $i$-th behavior mined
from domain $d$, and $n_d$ for the number of behaviors mined from domain $d$.

At inference, we condition generation on a small subset of behaviors selected for relevance to
the current query $x$: given $e(x)$, a \emph{selection rule} $\mathrm{Sel}$ returns a set of
behaviors whose natural-language
descriptions are inserted into the prompt. In the \textit{online} setting we study in this work, new pairs of (query, mined-behavior) arrive sequentially as new samples are seen, and the organizing structure itself must be able to adapt and incorporate each one, and optionally, perform periodic reorganization. We formalize any such structure with these properties as three operations on a state $\Omega$, which represents the current memory organization:
\[
\mathrm{retrieve}: \bbR^D \times \Omega \to \Delta^{K}, \qquad
\mathrm{update}: \bbR^D \times \Omega \to \Omega, \qquad
\mathrm{refit}: \Omega \to \Omega
\]

Here, $\Delta^K$ is the simplex of weights over the set of components. If the mechanism has a flat structure, $K$ is simply the total number of behaviors (each behavior is treated as its own component); otherwise, if there is clustering, we take $K$ to be the number of currently active components. This allows us to consider different practical instantiations (organizing structures) under a consistent online setup, with these being the central three operations of the online TTT loop.

\paragraph{Starting Point: \textit{k}-means clustering.} The simplest instantiation partitions $\bbR^D$ into $K$ hard clusters with centroids
$\mu_1,\dots,\mu_K$, i.e.\ $\Omega = (\mu_1,\dots,\mu_K, n_1,\dots,n_K)$ where $n_k$ counts
points currently assigned to cluster $k$.
\[
\mathrm{retrieve}(e(x),\Omega)
     = \mathrm{softmax}_k\!\left(
         -\frac{\|e(x)-\mu_k\|^2}{2\sigma^2}
       \right), \qquad
\sigma^2 = \frac{1}{K}\sum_k \sigma_k^2,
\]
where $\sigma_k^2$ is the scalar cluster variance. Given a newly mined behavior with embedding $e_t$, the $\mathrm{update}(e_t, \Omega)$ operator allows for the mechanism to grow the number of clusters by checking whether $1-\max_j\cos(e_t,\mu_j)$ exceeds a threshold $\tau_{\mathrm{KM}}$. If so, it creates a new cluster with centroid $\mu_{K+1}=e_t$ and count $n_{K+1}=1$. Otherwise, it assigns the behavior to the nearest centroid $k^\star$ and updates, while every other centroid is unchanged:
 \[
     \mu_{k^\star}\leftarrow
     \frac{n_{k^\star}\mu_{k^\star}+e_t}{n_{k^\star}+1},
     \qquad
     n_{k^\star}\leftarrow n_{k^\star}+1
     \]

     The $\mathrm{refit}(\Omega)$ operator entails one full pass of Lloyd's algorithm: all accumulated points are reassigned to their nearest centroids, each centroid is recomputed as the mean of its assigned points, and any remaining empty clusters are removed.

This approach has a few key limitations. First, hard assignment assumes clear specialization of the components in natural language, while behaviors perhaps should actually lie in multiple clusters, and $\textit{k}$-means requires prescribing the number of clusters a priori for its initialization; we want this to be determined in a data-driven manner. Second, "growth" (in terms of the creation of new components) is not tied to the probability-based retrieval mechanism, which can lead to many singleton components or a few overloaded components. Also, $\textit{k}$-means remains flat and unable to express the notion of domain-specificity; we want a structure that can encode domain-awareness while allowing for behaviors to possibly be shared across domains, motivating a hierarchy.

\section{Bayesian Cheatsheet: Behavior Mixtures via HDP-GMM}\label{sec:method}

We develop the \textbf{Bayesian Cheatsheet}, a nonparametric instantiation of a structured cheatsheet using a Dirichlet Process Gaussian Mixture Model (DP-GMM) \citep{rasmussen2000} to model behavior embeddings and counts for each component. This serves as an analogue to the $k$-means approach, with probabilistically latent assignment in lieu of hard assignment, countably infinite components with probability-driven growth rather than a fixed number, and an extension to a Hierarchical DP-GMM for domain-aware modeling.
The detailed algorithm is included in Appendix \ref{app:algo}.

\subsection{Generative Mixtures with Dirichlet Processes}\label{sec:dp-mixture}

First consider a single domain. We draw a random probability measure:
\[
  G\sim\DP(\gamma,H),
  \qquad
  H=\NIW(\mu_0,\kappa_0,\Psi_0,\nu_0)
\]
with the NIW distribution being the conjugate base distribution over Gaussian parameters. G is a discrete random probability measure, and thus can be written as
\[
  G=\sum_{k=1}^{\infty}\pi_k\delta_{\phi_k},
  \qquad
  \phi_k=(\mu_k,\Sigma_k),
  \qquad
  \phi_k\overset{\mathrm{i.i.d.}}{\sim}H
\]
as many observations can share the same atomic value $\phi_k$. We model each behavior embedding $e_i$ as:
  \[
  z_i\mid\pi\sim\Cat(\pi),
  \qquad
  e_i\mid z_i=k\sim\Nor(\mu_k,\Sigma_k).
  \]

  Marginalizing over $G$ yields the Chinese Restaurant Process (CRP). Conditioned on the previous assignments, a new behavior is assigned to
  an existing component $k$ with probability proportional to its occupancy
  $n_k$, or instantiates a new component with probability proportional to
  $\gamma$. As a result, the number of components ($K$) is determined in a data-driven manner rather than fixed a priori.

  Given a mined corpus of behavior embeddings $X$, the $\mathrm{retrieve}(x)$ operation uses the posterior assignment probability:
    \begin{equation}
    \label{eq:single-domain-posterior}
  p(z_x=k\mid e(x),X)
  \propto
  n_k\,p(e(x)\mid X_k)
  \end{equation}

  where $X_k=\{e_i:z_i=k\}$ and $p(e(x)\mid X_k)$ is the posterior predictive density for component $k$. Correspondingly, the unnormalized score for spawning a new component is given by $\gamma\,p(e(x)\mid H)$. Under the NIW prior, the predictive densities are multivariate Student-$t$ distributions.

  Crucially, the posterior of each component depends only on three additive sufficient statistics:
  \begin{equation}
\label{eq:suffstats}
n_k = |\{i : z_i = k\}|, \qquad
S_k^{(1)} = \textstyle\sum_{i: z_i=k} e_i, \qquad
S_k^{(2)} = \textstyle\sum_{i: z_i=k} e_i e_i^\top
\end{equation}
  Thus, incorporating a newly discovered behavior at test-time requires only incrementing these three values, allowing for online adaptation without needing to refit the full mixture after each observation.

\subsection{Hierarchical Construction for Cross-Domain Structure}

A single DP-GMM fit on a pooled corpus of cross-domain problems and behaviors cannot express the prevalence of components at the domain-specific level. At the same time, fitting independent DP-GMMs per domain can result in redundancy and unnecessary overhead, preventing components from being shared. For instance, a behavior like "use modular arithmetic" could be useful for a number theory problem as well as for periodic boundary conditions in a physics problem.

A Hierarchical Dirichlet Process (HDP) \citep{teh2006} addresses this by drawing the atoms $\{\phi_k\}$ once at the global level, and letting each domain draw its own DP over them, with $\theta_{d,i}=(\mu_{d,i},\Sigma_{d,i})$:
\begin{align*}
  G_0 &\sim \DP(\gamma,H),\\
  G_d\mid G_0 &\sim \DP(\alpha_d,G_0),
      \hspace{2mm} \forall \hspace{0.5mm}d\in\cD,\\
  \theta_{d,i}\mid G_d &\sim G_d,\\
  e_{d,i}\mid\theta_{d,i}
  &\sim\Nor(\mu_{d,i},\Sigma_{d,i})
  \end{align*}

  Thus, the Gaussian parameters $\phi_k=(\mu_k,\Sigma_k)$ are shared
  across domains, while the mixing weights $\pi_d$ are domain specific. Marginalizing the global and domain-specific random measures yields the
  Chinese Restaurant Franchise (CRF) representation.

  As in the single-domain construction, let $z_{d,i}=k$ denote that
  behavior $(d,i)$ uses the shared atom $\phi_k$. Conditional on $\pi_d$,
  this assignment has probability
  $p(z_{d,i}=k\mid\pi_d)=\pi_{d,k}$. We distinguish the local occupancy of component $k$ in domain $d$, $n_{d,k} = |\{i:z_{d,i}=k\}|$, from its global occupancy $N_k = \sum_{d \in \cD} n_{d,k}$.
  The local counts determine the domain-specific mixing weights, whereas
  the posterior of the shared Gaussian component uses observations pooled
  across domains. The sufficient statistics (Equation~\ref{eq:suffstats}) generalize by using $N_k$ and
  summing $S_k^{(1)}$ and $S_k^{(2)}$ over every domain.

\subsection{Posterior Inference via Collapsed Gibbs Sampling}\label{sec:gibbs}

To fit the HDP-GMM, we use a collapsed Gibbs sampler \citep{neal2000} in the CRF representation, leveraging NIW-Gaussian conjugacy in marginalizing out the Gaussian parameters $\phi_k$. We also integrate out the random measures, retaining local groups and their mappings to shared global components \citep{teh2006}. Marginalizing over the local-group choice gives the assignment probability for an existing component:
\begin{equation}
\label{eq:hierarchical-assignment}
p(z_{d,i}=k\mid\cdot)
\propto
\left(n_{d,k}^{-i}+\alpha_d\widehat\beta_k^{-i}\right)
p_k^{-(d,i)}(e_{d,i}),
\qquad k=1,\ldots,K
\end{equation}
Here, $n_{d,k}^{-i}$ counts behaviors assigned to component $k$ within domain $d$, $\widehat\beta_k^{-i}$ is the global predictive mass derived from the group-to-component counts, and $p_k^{-(d,i)}$ is the Student-$t$ posterior predictive computed from behaviors pooled across domains -- all excluding observation $(d,i)$.

Relative to the single-domain posterior rule (\ref{eq:single-domain-posterior}), the hierarchical structure lets the memory reflect both domain-specific patterns as well as cross-domain reuse. The local count $n_{d,k}^{-i}$ concentrates observations onto prevalent components that are already common within domain $d$, following the "rich-get-richer" effect associated with DPs. The $\alpha_d\widehat\beta_k^{-i}$ term allows a domain to draw on components learned across the corpus through the mixing weights. The predictive density weights this by how well the behavior's embedding fits the component, so assignment depends on both the behavior itself and the frequency of its component. Overall, this allows the memory to share existing reasoning patterns across domains while also accommodating additional components as behaviors accumulate.

A complete step of partially collapsed Gibbs resamples every $z_{d,i}$ in random order and recomputes both the local counts and the globally pooled sufficient statistics. Offline fitting of the HDP-GMM repeats this procedure for a fixed iteration count. Further mathematical details on the two-level sampling procedure for our collapsed Gibbs sampling process are included in Appendix~\ref{app:gibbs}.

\subsection{Test-Time Adaptation}\label{sec:method-ttt}

\paragraph{Retrieval via the posterior predictive.}

In the offline phase, the embedding used is the behavior's own embedding (applying $\cE$), and we apply PCA for efficiency in fitting the mixture model (described in Appendix~\ref{app:impl-hparams}). At inference time, $e(x)$ is approximated with a small learned projector $f_{\theta}$ to map raw query embeddings into the same space (see Appendix~\ref{app:impl}). With domain-conditioning, the query's domain is inferred using a nearest-centroid classifier, where $c_d$ is the centroid of domain $d$'s data:
  \[
  d_x = \widehat d(x)
  =
  \argmin_{d\in\cD}
  \operatorname{dist}\!\left(e(x),c_d\right)
  \]
    Then $\mathrm{retrieve}(e(x), \Omega)$   combines these domain-specific  mixture weights with
  the globally shared component predictives, so domains can share the same components but retrieve them with different probabilities.
  The normalized weight for each component under the Student-$t$ posterior predictive is:
  \begin{equation}
  \label{eq:selection}
  w_{d_x,k}(e(x))
  =
  \frac{
  \widehat\pi_{d_x,k}\,p_k(e(x))
  }{
  \sum_{j=1}^{K}
  \widehat\pi_{d_x,j}\,p_j(e(x))
  }
  \end{equation}

We consider component selection either by taking the top-$k'$ components by weight or choosing every component whose weight exceeds a threshold $\tau$. Once components are selected, the medoid element is returned for each component, treated as its representative element.

\paragraph{Online updates and new component creation.} After generating a response to query $x_t$, the model mines a new behavior by self-reflection on its reasoning, with embedding $e_t$ and domain $d_t = \widehat d(x_t)$. A discrete CRF partition produces the mixing weights, while responsibility-weighted Gaussian statistics determine the online densities; the predictive probability of a new component is:
\begin{equation}
  \label{eq:novelty}
  \rho_{\mathrm{new}}(e_t)
  =
  \frac{
  \widehat\pi_{d_t,\mathrm{new}}p_0(e_t)
  }{
  \widehat\pi_{d_t,\mathrm{new}}p_0(e_t)
  +
  \sum_{k=1}^{K}\widehat\pi_{d_t,k}\,p_k(e_t)
  }
  \end{equation}
  Inspired by \citet{kulis2012dpmeans}, we use thresholding: if $\rho_{\mathrm{new}}(e_t)$ exceeds
  $\tau_{\mathrm{new}}$, we spawn a new globally shared component, initialize its Gaussian sufficient statistics using $e_t$, and create a local group containing this behavior. Otherwise, it is softly added to existing components with $w_{d_t,k}(e_t) >
    10^{-6}$, weighted accordingly. The effective Gaussian sufficient statistics for $\mathrm{update}(e_t, \Omega)$ are:
  \begin{align}
    \widetilde N_k &\leftarrow \widetilde N_k+w_{d_t,k}(e_t), &
    \widetilde S_k^{(1)} &\leftarrow \widetilde S_k^{(1)}+w_{d_t,k}(e_t)e_t,&
    \widetilde S_k^{(2)} &\leftarrow \widetilde S_k^{(2)}+w_{d_t,k}(e_t)e_te_t^\top
    \end{align}

    We also separately store the hard assignment $z_t=\argmax_k
    w_{d_t,k}(e_t)$ and sample its local group assignment conditional on  $z_t$, incrementing the local group element count and $n_{d_t, z_t}$. If a new local group is created, we also increment $M_{z_t}$, the global group count.

\textbf{Periodic refitting.} Every $g$ samples (the refit interval), we perform $\mathrm{refit}(\Omega)$ by rebuilding Gaussian statistics from the stored hard assignments and performing a single collapsed Gibbs step over all behaviors, keeping domain labels fixed. We then greedily merge components whose symmetric Gaussian KL-divergence is $< \epsilon$, updating group mappings, counts, and statistics. This allows for reorganization to reconcile drift that may be introduced by performing many soft updates over time.

\section{Results}\label{sec:results}

\subsection{Experimental Setup}\label{sec:exp-setup}

\paragraph{Models.} All experiments are performed with the Qwen3-32B \citep{yang2025qwen3} and Granite-4.2-30B \citep{ibm_granite_4_2_30b} language models, with thinking/reasoning modes enabled. Allowing language models to leverage their reasoning abilities to the fullest ensures that retrieved reasoning behaviors from cheatsheets can be reflected in their natural language reasoning traces.

\paragraph{Datasets.} We evaluate on three reasoning-intensive benchmarks: AIME'25 \citep{aime2025}, Omni-MATH\footnote{Specifically, we consider a hard subset consisting only of problems with $\mathrm{difficulty} \geq 7.0$.} \citep{gao2025omnimath}, and PhysReason\footnote{We consider solely the questions which do not require an image (i.e. text-only), given that the open-weight language models we consider do not have multimodal capabilities, yielding 226 questions (mini has 31).} \citep{zhang2025physreason}, averaged over 3 seeds. Our behavior corpus (existing memory) was mined offline from NuminaMath-CoT \citep{li2024numinamath} and the Camel Physics \citep{li2023camel} datasets; the respective language model solves each problem and produces a self-reflection of the behaviors that were useful in solving the problem, regardless of the answers' correctness. The corpus was mined separately for each language model, providing an on-policy reasoning bank. Further details on the behavior corpus are included in Appendix \ref{app:behavior-corpus}.

\paragraph{Baselines.} We consider a number of baselines, evaluated both with and without the pre-mined memory, with the key differences being the retrieval and adaptation mechanisms. This includes: 1. random behavior selection of $k'$ elements; 2. Metacognitive Reuse \citep{didolkar2025metacognitive}, which involves flat cosine-similarity retrieval over the behavior library (top-$k'$) in embedding space; and 3. direct behavior retrieval, performing flat retrieval in the $f_{\theta}$-projected space rather than embedding space. For the dynamic (test-time adaptation) setting, we consider the Dynamic Cheatsheet \citep{suzgun2026cheatsheet} method (both DC-Cu and DC-RS), which grows the library as it solves, with up to \textrm{max\_new}=3 elements added per sample, and has the ability to rewrite and refine the cheatsheet purely in natural language (as it has a flat structure). We also consider the ACE method \citep{zhang2026ace} which, on top of Dynamic Cheatsheet, introduces iterative refinement through a reflector as well as editing and deduplication. In practice, we use $k'=5$ for all experimental results.

\paragraph{Methods.} We evaluate the Bayesian Cheatsheet (BC) as described in Section \ref{sec:method}, as well as $k$-means clustering as introduced in Section \ref{sec:structured-retrieval}. We also include BC-centroids, which selects the $k'$ closest centroids over the HDP-GMM-fitted components, rather than posterior predictive selection. This allows us to isolate the role of 1. structure (these mechanisms against the baselines), 2. adaptation (HDP-GMM vs $k$-means), 3. flexible assignment (hard $k$-means vs soft $k$-means and HDP-GMM), and 4. component selection (posterior predictive vs. nearest-centroid top-$k'$). To bring the $k$-means algorithms to a level playing field with the HDP-GMM mechanism, we introduce adaptive versions with online updates for each, along with analogues for the Gibbs refit. Hard $k$-means assigns each new behavior to its nearest centroid and updates that centroid, while refitting is just Lloyd reassignment. For soft $k$-means (an isotropic generative variant), the online updates fold each new behavior into every cluster with weight above $10^{-6}$, with the refitting being a responsibility-weighted soft EM procedure. To standardize across methods, we take $K$ (number of initial clusters for $k$-means methods) to be the same number of components yielded in fitting the HDP-GMM, yielding 64 clusters (Qwen), and 62 clusters (Granite) -- we include a list of  labels for each component in Appendix \ref{app:component-labels}.

\subsection{Bayesian Cheatsheets Improve Metacognitive Reasoning}

\begin{figure}[t]
\centering
\includegraphics[width=\linewidth]{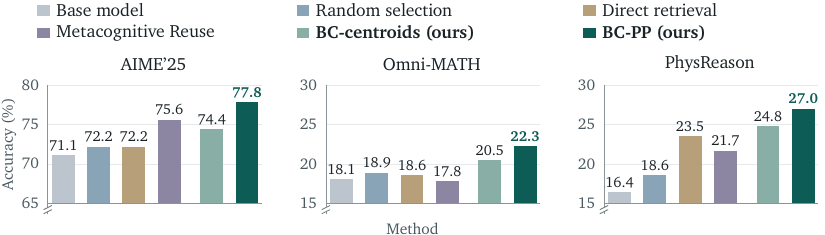}
\captionsetup{skip=4pt}
\caption{Static retrieval with Qwen3-32B using the mined behavior corpus. Bayesian Cheatsheet -- specifically, our posterior predictive strategy (BC-PP) -- achieves the strongest performance across the board, while our top-$k'$ nearest-centroid variant also performs well relative to the baselines.}
\label{fig:static-retrieval}
\vspace{-8pt} \end{figure}

Figure~\ref{fig:static-retrieval} and Appendix Table~\ref{tab:baselines} show that BC-PP improves over Metacognitive
Reuse by +2.2--5.3 points for Qwen and +1.3--3.2 for Granite across all three benchmarks. While BC-PP generally outperforms BC-centroids (with the exception of AIME'25 and a tie for PhysReason on Granite), the latter outperforms the baselines in nearly all settings. Since these methods have access to the same offline-mined corpus for each respective language model, our gains indicate that memory structure and component selection are vital. Furthermore, the results with BC-PP indicate that there is additional value yielded through the posterior-predictive selection mechanism beyond organizing behaviors into components. Thus, effective reasoning strategy reuse depends not only on the availability of useful behaviors, but also
on how the memory makes them accessible for a given query.

\subsection{Bayesian Cheatsheet Enables Online Adaptation}

We next evaluate Bayesian Cheatsheet in the online adaptation setting, where we denote it as BC-TTT. As described in Section \ref{sec:method-ttt}, we allow for new behaviors to be generated through self-reflection and subsequently either added to existing components or used to seed new components that are spawned.

Our results in Table~\ref{tab:main-result} show that online updates improve Bayesian Cheatsheet over
static retrieval across both models and all benchmarks, even without
periodic refitting. Our comparison against the $k$-means methods (described in Section \ref{sec:structured-retrieval}) demonstrates that 1) soft assignment can be valuable, as soft $k$-means matches or outperforms the hard assignment variant on AIME'25 and PhysReason, and 2) the HDP-GMM structure helps with domain-conditioning in retrieval as well as principled growth in the online setting.
Our comparisons against Dynamic Cheatsheet and ACE -- methods which also incorporate experience through online memory updates, albeit in a flat structure -- are particularly relevant. The gains of BC-TTT over these methods suggest that component-based online adaptation is effective; while allowing the organization of the cheatsheet to adapt based on the novelty of the behaviors encountered can help, it could also have tradeoffs, as seen with Granite slightly degrading on AIME'25 and Omni-MATH, but still continuing to outperform all other methods.

  \begin{table}[t]
  \centering
  \captionsetup{position=top,skip=\baselineskip}
  \caption{Static prompting vs.\ online adaptation, with and without periodic
  refitting ($g=50$ for OM and PR, $g=10$ for AIME'25), for Qwen3-32B and Granite-4.2-30B. The static version of DC-RS is
Metacognitive Reuse (in Table~\ref{tab:baselines}). DC-Cu and DC-RS involve "curation" where the cheatsheet can be refined (does not have a separate refit), while ACE similarly uses a curator for adaptation and does not have a static equivalent. We denote these entries with a "--". Bayesian Cheatsheet (BC-TTT) improves over existing methods across models and
  benchmarks, both with and without refit.}
  \label{tab:main-result}

  \begingroup
  \footnotesize
  \setlength{\tabcolsep}{2pt}
  \renewcommand{\arraystretch}{1.08}

  \begin{tabular*}{\linewidth}{@{\extracolsep{\fill}}ll*{6}{c}@{}}
  \toprule
  & & \multicolumn{3}{c}{Qwen3-32B}
    & \multicolumn{3}{c}{Granite-4.2-30B} \\
  \cmidrule(lr){3-5}\cmidrule(lr){6-8}
  Method & Condition
    & AIME'25 & \shortstack{OM} & \shortstack{PR}
    & AIME'25 & \shortstack{OM} & \shortstack{PR} \\
  \midrule

  \multirow{3}{*}{Hard $k$-means}
   & Static             & 73.3 & 20.3 & 18.7 & 75.6 & 27.0 & 18.3 \\
   & Online, no refit    & 75.6 & 24.5 & 23.5 & 77.8 & 28.6 & 20.4 \\
   & Online, with refit  & 74.4 & 24.8 & 24.8 & 76.7 & 31.6 & 24.6 \\
  \midrule

  \multirow{3}{*}{Soft $k$-means}
   & Static             & 73.3 & 21.6 & 23.6 & 75.6 & 28.4 & 23.2 \\
   & Online, no refit    & 75.6 & 23.4 & 28.5 & 78.9 & 26.9 & 22.0 \\
   & Online, with refit  & 76.7 & 23.1 & 29.1 & 77.8 & 29.8
                        & \underline{27.4} \\
  \midrule

  \multirow{3}{*}{DC-Cu}
   & Static             & 73.3 & 17.9 & 18.1 & 74.4 & 26.8 & 18.1 \\
   & Online, curation   & 75.6 & 21.8 & 24.6 & 77.8 & 28.3 & 22.4 \\
   & Online, with refit  & --   & --   & --   & --   & --   & --   \\
  \midrule

  \multirow{3}{*}{DC-RS}
   & Static             & 75.6 & 17.8 & 21.7 & 77.8 & 28.4 & 22.6 \\
   & Online, curation   & 78.9 & 21.0 & 25.7 & 81.1 & 30.2 & 23.0 \\
   & Online, with refit  & --   & --   & --   & --   & --   & --   \\
  \midrule

  \multirow{3}{*}{ACE}
   & Static             & --   & --   & --   & --   & --   & --   \\
   & Online, curator    & 78.9 & 23.2 & 26.0 & 81.1 & 29.3 & 25.1 \\
   & Online, with refit  & --   & --   & --   & --   & --   & --   \\
  \midrule

  \multirow{3}{*}{\textbf{BC-TTT (ours)}}
   & Static
   & 77.8 & 22.3 & 27.0 & 80.0 & 31.6 & 23.9 \\
   & Online, no refit
   & \underline{81.1}
   & \underline{26.4}
   & \underline{30.1}
   & \textbf{85.6}
   & \textbf{34.6}
   & 26.8 \\
   & Online, with refit
   & \textbf{82.2}
   & \textbf{26.6}
   & \textbf{31.3}
   & \underline{84.4}
   & \underline{34.3}
   & \textbf{30.7} \\
  \bottomrule
  \end{tabular*}
  \endgroup
  \end{table}

\paragraph{Periodic Refitting of Components.}

We quantify how much a refit changes the behavior-component assignment with a maximum-overlap bipartite matching. Let $\ell_{\mathrm{before}}, \ell_{\mathrm{after}}$ be the cluster labels assigned to the $n$ accumulated points immediately before and after a refit (which may consist of different label sets and active label counts). Consider the contingency table:
\begin{equation}
\label{eq:contingency}
M[b,a] = \big|\{i \in [n] : \ell_{\mathrm{before}}(i) = b,\; \ell_{\mathrm{after}}(i) = a\}\big|
\end{equation}
We want to find the label correspondence maximizing total overlap over one-to-one matchings $\sigma$:
\begin{equation}
\label{eq:hungarian}
\sigma^\star = \argmax_{\sigma} \sum_b M[b, \sigma(b)]
\end{equation}
This is an assignment problem, which is solvable with the Hungarian algorithm. A point is counted as a reassignment only if its label in $\ell_{\mathrm{after}}$ differs from the matched image of its label in $\ell_{\mathrm{before}}$. A pure relabeling should receive zero churn, and we are only counting the points that change clusters:
\begin{equation}
\label{eq:churn}
\text{\emph{churn}} = \frac{1}{n}\sum_{i=1}^n \mathds{1}\!\left[\ell_{\mathrm{after}}(i) \neq \sigma^\star(\ell_{\mathrm{before}}(i))\right],
\end{equation}

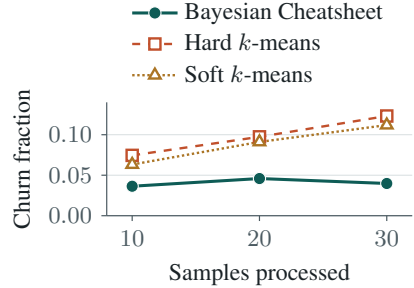
\begin{wrapfigure}{r}{0.40\textwidth}
\vspace{-\intextsep}
\centering
\captionsetup{font=small,skip=4pt}
\pgfplotsset{
  churn axis/.style={
    scale only axis,
    width=\dimexpr\linewidth-44pt\relax, height=1.50cm,
    xmin=8, xmax=32, ymin=0, ymax=0.14,
    xtick={10,20,30}, ytick={0,0.05,0.10},
    scaled y ticks=false,
    yticklabel style={
      /pgf/number format/fixed,
      /pgf/number format/precision=2,
      /pgf/number format/fixed zerofill
    },
    xlabel={Samples processed},
    label style={font=\small, text=bmink},
    tick label style={font=\footnotesize, text=bminkgrey},
    title style={font=\small\mdseries, text=bmink, yshift=-1mm},
    axis lines=left,
    axis line style={draw=bminkgrey, line width=0.45pt, -},
    tick align=outside, tick pos=left,
    major tick length=2pt,
    tick style={draw=bminkgrey, line width=0.45pt},
    ymajorgrids=true,
    grid style={draw=bmline, line width=0.35pt},
    legend columns=1,
    legend style={
      font=\footnotesize, text=bmink,
      draw=none, fill=none,
      at={(0.5,1.05)}, anchor=south,
      cells={anchor=west}, row sep=0pt
    },
    legend image post style={mark repeat=2,mark phase=2}
  },
  churn hdp/.style={
    color=bmteal, line width=1.15pt,
    mark=*, mark size=2.3pt,
    mark options={solid, fill=bmteal, draw=white, line width=0.4pt}
  },
  churn hard/.style={
    color=bmterra, line width=0.9pt, dashed,
    mark=square*, mark size=2.1pt,
    mark options={solid, fill=white, draw=bmterra, line width=0.9pt}
  },
  churn soft/.style={
    color=bmochre!85!black, line width=0.9pt, densely dotted,
    mark=triangle*, mark size=2.5pt,
    mark options={solid, fill=white, draw=bmochre!85!black, line width=0.9pt}
  }
}
\begin{tikzpicture}
\begin{axis}[
  churn axis,
  ylabel={Churn fraction}
]
\addplot[churn hdp] coordinates {(10,0.0364) (20,0.0459) (30,0.0398)};
\addlegendentry{Bayesian Cheatsheet}
\addplot[churn hard] coordinates {(10,0.0745) (20,0.0974) (30,0.1231)};
\addlegendentry{Hard $k$-means}
\addplot[churn soft] coordinates {(10,0.0632) (20,0.0911) (30,0.1119)};
\addlegendentry{Soft $k$-means}
\end{axis}
\end{tikzpicture}
\caption{Assignment churn (Qwen3-32B) on PhysReason-mini; refits every 10 queries.}
\label{fig:churn}
\end{wrapfigure}

Figure~\ref{fig:churn} shows that Bayesian Cheatsheet's assignment churn remains
near 4\% across refits on PhysReason-mini. At the third refit, churn
is 3.98\%, compared to 12.31\% for hard $k$-means and 11.19\% for soft
$k$-means. The gap becomes wider over the refit rounds, as both $k$-means methods have increasingly larger changes in component membership. As a result, this indicates that Bayesian Cheatsheet's online updates preserve more of the existing organization through refitting while component parameters and domain-specific weights still evolve as new behaviors are incorporated. Thus, churn is a measure of the stability of memory's organization, and while BC is still subject to changing through online adaptation, it is quite stable.

\par
\ifnum\value{WF@wrappedlines}>2
  \vspace{\dimexpr\value{WF@wrappedlines}\baselineskip-2\baselineskip\relax}
\fi
\WFclear

\paragraph{Cold-Start Memory Accumulation.} Thus far, we have considered the setting where an offline-mined behavior corpus is available to the language model as memory in metacognitive reasoning. What if the language model were instead given a blank slate and tasked with learning the cheatsheet on-the-fly, purely through self-reflection? We evaluate our method (an empty memory initialized with an NIW prior) as well as ACE and DC-Cu (in their cold-start constructions), to determine whether Bayesian Cheatsheet is dependent on a warm-started memory module.

\begin{table}[h]
  \centering
  \captionsetup{position=top,skip=\baselineskip}
  \caption{Accuracy on Omni-MATH ($n=916$) and PhysReason ($n=226$)
  with no pre-mined behavior corpus. Memory-based methods start from an
  empty memory and accumulate behaviors through self-reflection at test
  time. Bayesian Cheatsheets use a periodic refit interval of $g=50$.}
  \label{tab:cold-start}

  \begingroup
  \footnotesize
  \setlength{\tabcolsep}{1pt}
  \renewcommand{\arraystretch}{1.025}

  \begin{tabular*}{0.88\linewidth}{@{\extracolsep{\fill}}l*{4}{c}@{}}
  \toprule
  & \multicolumn{2}{c}{Qwen3-32B}
  & \multicolumn{2}{c}{Granite-4.2-30B} \\
  \cmidrule(lr){2-3}\cmidrule(lr){4-5}
  Method & OM & PR & OM & PR \\
  \midrule
  Base model (no memory)
  & 18.1 & 16.4
  & 24.8 & \underline{16.4} \\
  \midrule
  DC-Cu, cold-start
  & \underline{22.5} & 15.9
  & 25.5 & 15.3 \\
  ACE, cold-start
  & 21.0 & \underline{17.7}
  & \underline{27.9} & \underline{16.4}  \\
  \midrule
  \textbf{BC-TTT, cold-start (ours)}
  & \textbf{24.1} & \textbf{20.5}
  & \textbf{31.9} & \textbf{18.1} \\
  \bottomrule
  \end{tabular*}
  \endgroup
  \end{table}

  Our findings in Table \ref{tab:cold-start} with Omni-MATH and PhysReason show that even without an offline-mined corpus, Bayesian Cheatsheet outperforms cold-start baselines for both language models and clearly improves over the memory-free setting. PhysReason seems particularly difficult to improve on in the cold-start setting (DC-Cu regresses), suggesting that physics benefits from more extensive guidance; nonetheless, BC-TTT still achieves gains. This demonstrates that Bayesian Cheatsheet can construct useful behavior memory purely from test-time experience, aided by its dynamic reorganization capabilities with periodic refitting, achieving effective adaptation to new data.

\section{Discussion}

We introduce the \textit{Bayesian Cheatsheet}, a structured memory module for language models which uses Dirichlet Process Gaussian Mixture Models for component structure. We leverage this mechanism to organize reasoning strategies for inference-time retrieval, and demonstrate that this mechanism exhibits desirable properties in continual test-time adaptation while achieving clear gains in performance across reasoning benchmarks. Bayesian Cheatsheets offer a potential mechanism for offline domain adaptation  through their hierarchical design; new behaviors and mixing weights can be learned for unseen domains, while also inferring shared cross-domain structure. In this work, we focus on single-turn problem solving with small-to-medium-size non-agentic language models ($\approx 30B$ parameters); while agentic and multi-turn benchmarks are currently beyond our scope, they offer practical potential for iterative retrieval from the Bayesian Cheatsheet. Furthermore, the refitting and retrieval costs eventually grow over indefinitely long evaluation streams, suggesting the need for principled \textit{discarding} mechanisms to selectively retain behaviors. Overall, our findings highlight the potential for Bayesian-inspired methods to yield value for test-time adaptation of language models, given the setting's analogues to online learning, offering an effective, statistically-grounded mechanism.

\newpage
\appendix
\section{Additional Related Work}\label{app:related-work}

\paragraph{LLM reasoning methods.}
In-context learning provides language models with the ability to infer a task based on demonstrations, without updating their parameters \citep{brown2020fewshot}, a behavior that has an implicit Bayesian interpretation \citep{xie2022iclbayes}. Reasoning-oriented prompting makes intermediate computations available in natural language in the context through scratchpads \citep{nye2021scratchpads} and chain-of-thought \citep{wei2022cot}, which enables approaches that decompose problems into simpler subproblems \citep{zhou2023leasttomost}, aggregate multiple reasoning paths through self-consistency \citep{wang2023selfconsistency}, or search over alternative intermediate steps \citep{yao2023tot}.

Self-reflection extends this notion by allowing models to critique and revise their own solutions. Self-Refine alternates between generation, feedback, and revision \citep{madaan2023selfrefine}, while Reflexion retains verbal feedback in episodic memory for subsequent attempts \citep{shinn2023reflexion}. Related work on metacognition studies whether models can extract generalizable lessons from problem solving \citep{didolkar2024metacognitive}, although some evaluations of intrinsic self-correction show that revisions can fail to improve reasoning without external feedback \citep{huang2024selfcorrect}. These behaviors can also be learned through training, as in STaR's fine-tuning on self-generated rationales that lead to correct answers \citep{zelikman2022star}, training to self-correct
\citep{kumar2024score,ramji2025latent}, and Experiential Reinforcement Learning's use of feedback-guided revisions \citep{shi2026erl}.

\paragraph{Bayesian modeling for natural language.}
Bayesian nonparametric models have been used in language tasks to learn latent structures where size is not fixed a priori, ranging from word segmentation \citep{goldwater2006segmentation} to the reusable grammar subtrees of adaptor grammars \citep{johnson2007adaptor}. The hierarchical Dirichlet process \citep{teh2006}
is particularly useful in topic modeling, allowing for topics to be shared across documents while having document-specific proportions, while hierarchical Pitman--Yor language models \citep{teh2006pyp} are useful for different-length contexts. Related hierarchical sharing appears in domain adaptation, where \citet{finkel2009hierarchical} link domain-specific model parameters through a shared Gaussian prior, which balances cross-domain guidance with domain-specific observations.
For continuous representations, Gaussian LDA \citep{das2015gaussianlda} models word embeddings with Gaussian topic distributions with a fixed topic inventory. By contrast, infinite Gaussian mixtures, as introduced in
\citep{rasmussen2000}, should allow the number of occupied components to grow with the data. Bayesian inference has also been used with modern language models in uncertainty quantification and guided generation. Laplace-LoRA approximates a posterior over adaptation parameters for improved calibration \citep{yang2024laplacelora}, while \citet{hu2024amortizing} use amortized posterior inference for constrained generation and implicit reasoning.

\paragraph{Test-time adaptation and continual learning (continued).}
SEAL \citep{zweiger2025seal} learns to generate
synthetic training data and the updates to adapt the language model (i.e. learning how to adapt itself). In architectures such as Titans \citep{behrouz2025titans} and ATLAS \citep{behrouz2025atlas}, adaptation instead takes place through a neural memory that learns from the incoming context. For agents, AutoHarness \citep{lou2026autoharness} uses environment feedback to synthesize and refine code that constrains an LLM's actions, while Hyperagents \citep{zhang2026hyperagents} make both the task-solving program and the procedure for modifying it editable. Harness Continual Learning \citep{kang2026harness} treats prompts, memory, skills, and routing as a persistent, jointly updated state, evaluating candidate changes to improve on the current task while retaining earlier capabilities before deployment.

\paragraph{Memory and retrieval-augmented reasoning with LLMs (continued).}

Nonparametric retrieval allows for new information to be incorporated without retraining the language model; nearest-neighbor language models can do this by changing the retrieval datastore \citep{khandelwal2020knnlm}. Hierarchical retrieval methods organize information at multiple levels of abstraction: for example, RAPTOR recursively clusters and summarizes document passages into a retrieval tree \citep{sarthi2024raptor}, while MemTree incrementally builds a hierarchy of semantic summaries as observations arrive \citep{rezazadeh2024memtree}. Buffer of Thoughts maintains reusable thought templates as procedural memory \citep{yang2024bot}, and Memp distills trajectories into both detailed instructions and higher-level abstractions that can be revised over time with experience \citep{fang2026memp}. In MemRL, the utility of stored experiences is learned from feedback, so that retrieval accounts for their usefulness \citep{zhang2026memrl}. In ALMA, memory organization is a learned goal, where executable designs specifying what to store, how to retrieve, and how to update (the three key operations we defined in Section \ref{sec:structured-retrieval}) are searched over \citep{xiong2026alma}.

\section{Bayesian Cheatsheet Algorithm}\label{app:algo}

The memory (modeled by $\Omega$) stores records of behaviors and their embeddings, as well as domain, component, and local group assignments, local and global CRF counts, Gaussian sufficient statistics, and the medoid elements for retrieval.

\begin{algorithm}[htbp]
\caption{Bayesian Cheatsheet for Test-Time Adaptation (with periodic refitting)}
\label{alg:ttt}
\small
\begin{algorithmic}[1]
\Require Initial memory $\Omega_0$, NIW prior $H$, concentrations $\gamma,\{\alpha_d\}$, queries $x_{1:T}$
\Require Retrieval budget $k'$, novelty threshold $\tau_{\mathrm{new}}$, refit interval $g$ (or disabled), merge threshold $\epsilon$
\State Set $\Omega\gets\Omega_0$; compute component predictives, CRF weights, and medoids
\For{$t = 1, 2, \dots, T$}
  \State Embed query as $e(x_t)$ and infer $d_t=\widehat d(x_t)$; use the sole domain in cold start
  \State Set the retrieved behavior set $\mathcal R_t\gets\varnothing$
  \If{memory is nonempty}
    \State Select up to $k'$ highest-weight components  via $w_{d_t,k}(e(x_t))$ (Eq.~\ref{eq:selection})
    \State Set $\mathcal R_t$ to one medoid behavior per selected component
  \EndIf
  \State Generate the response using $\mathcal R_t$, then mine a behavior $b_t$ and compute its embedding $e_t$
  \State Set $\rho_{\mathrm{new}}(e_t)=1$ if memory is empty; otherwise evaluate Eq.~\ref{eq:novelty} on the current memory
  \If{memory is empty or $\rho_{\mathrm{new}}(e_t)>\tau_{\mathrm{new}}$}
    \State Create a global component $z_t$ with zero CRF counts; set $\ell_t\gets\mathrm{new}$
    \State $(\widetilde N_{z_t},\widetilde S_{z_t}^{(1)},\widetilde S_{z_t}^{(2)})\gets(1,e_t,e_te_t^\top)$
  \Else
    \State Cache $w_{d_t,k}(e_t)$ for all existing $k$ using Eq.~\ref{eq:selection}, before changing $\Omega$
    \For{each $k$ with $w_{d_t,k}(e_t)>10^{-6}$}
      \State $(\widetilde N_k,\widetilde S_k^{(1)},\widetilde S_k^{(2)})\gets(\widetilde N_k,\widetilde S_k^{(1)},\widetilde S_k^{(2)})+w_{d_t,k}(e_t)(1,e_t,e_te_t^\top)$
    \EndFor
    \State Set $z_t\gets\argmax_k w_{d_t,k}(e_t)$ using the cached weights
   \State Assign weight $n_{d_t,\ell}$ to each local group $\ell$ in $d_t$
     mapped to $z_t$
     \State Include a new group with weight $\alpha_{d_t}\widehat\beta_{z_t}$;
     normalize and sample $\ell_t$
  \EndIf
  \If{$\ell_t=\mathrm{new}$}
    \State Create an empty local group $\ell_t$ mapped to $z_t$; increment $m_{d_t,z_t}$ and $M_{z_t}$
  \EndIf
  \State Increment integer occupancies $n_{d_t,\ell_t}$ and $n_{d_t,z_t}$; store $(b_t,d_t,e_t,z_t,\ell_t)$ once
  \If{refitting enabled and $t\bmod g=0$}
    \State Rebuild Gaussian statistics from all stored hard assignments and refresh NIW predictives
    \State Run one random-order collapsed Gibbs step with fixed domains (Appendix~\ref{app:gibbs})
    \State Greedily merge pairs with symmetric Gaussian KL $<\epsilon$; update $\Omega$ and predictives at each merge
  \EndIf
  \State Refresh NIW predictives, all CRF mixing weights (Appendix~\ref{app:gibbs}), and medoids
\EndFor
\end{algorithmic}
\end{algorithm}

\section{Collapsed Gibbs Sampling}\label{app:gibbs}

This section expands on the collapsed Gibbs sampler in Section~\ref{sec:gibbs}, which organizes behavior embeddings into shared components while preserving domain-specific use patterns. We first obtain each component's posterior predictive from its pooled sufficient statistics, then describe the CRF assignment updates and the resulting domain-specific retrieval weights.

Given the global sufficient statistics for component $k$, the posterior over $\phi_k$ is $\NIW(\mu_k',\kappa_k',\Psi_k',\nu_k')$:
  \begin{align*}
  \kappa_k' &= \kappa_0+N_k,
  \qquad
  \nu_k' = \nu_0+N_k,\\
  \mu_k'
  &=
  \frac{\kappa_0\mu_0+S_k^{(1)}}{\kappa_k'}, \qquad
  \Psi_k'
  =
  \Psi_0+
  \left(S_k^{(2)}-N_k\bar e_k\bar e_k^\top\right)
  +
  \frac{\kappa_0N_k}{\kappa_k'}
  (\bar e_k-\mu_0)(\bar e_k-\mu_0)^\top
  \end{align*}

  Here, $\bar e_k = \frac{S_k^{(1)}}{N_k}$ for $N_k > 0$.   Integrating $\phi_k$ out of the Gaussian likelihood gives a multivariate Student-$t$ posterior predictive:
  \begin{equation}
  \label{eq:studentt}
  p_k(e)
  \equiv
  p(e\mid X_k)
  =
  t_{\nu_k'-D+1}\!\left(
  e
  \;\middle|\;
  \mu_k',
  \frac{\kappa_k'+1}
  {\kappa_k'(\nu_k'-D+1)}
  \Psi_k'
  \right)
  \end{equation}

For an empty component, the sufficient statistics are zero, so the posterior parameters reduce to $(\mu_0,\kappa_0,\Psi_0,\nu_0)$. Then, integrating over this NIW prior yields the prior predictive, which we denote $p_0(e)=p(e\mid H)$.

When resampling observation $(d,i)$, we use a two-level assignment/partition structure based on \cite{teh2006}. We maintain an auxiliary partition of the behavior observations within each domain $d$ into local groups, each of which is mapped to a global component, of which there are $K$ total. Multiple groups may map to the same component, inducing component-sharing across domains. Let $\ell_{d,i}$ denote the local-group assignment for sample $(d,i)$, so $n_{d,\ell}^{-i}$ is the count of behaviors already in group $\ell$ excluding $(d,i)$, $k_{\ell}$ is the component that group $\ell$ maps to, and $M_k^{-i}=\sum\limits_{d'\in\cD} m_{d',k}^{-i}$ is the total number of groups across all domains mapped to component $k$, excluding $(d,i)$.

A first categorical draw is done to assign $\ell_{d,i}$ over the set of existing local groups as well as a potential new group.
 \begin{equation}
\label{eq:gibbs-conditional}
p(\ell_{d,i}=\ell\mid\cdot)\propto
\begin{cases}
n_{d,\ell}^{-i} \, p_{k_\ell}^{-(d,i)}(e_{d,i}), \hspace{45mm}\ell\text{ is an existing local group}\\[4pt]
\frac{\alpha_d}
  {\gamma + \sum_{j=1}^{K}M_{j}^{-i}}
  \left(
  \sum_{k=1}^{K}
  M_k^{-i}p_k^{-(d,i)}(e_{d,i})
  +
  \gamma p_0(e_{d,i})
  \right), \hspace{5mm} \ell=\mathrm{new}
\end{cases}
\end{equation}

 Marginalizing over local group choice gives the equivalent component-level assignment score, for $\widehat \beta_{k}^{-i} = \frac{M_k^{-i}}
  {\gamma+\sum_{j=1}^{K}M_{j}^{-i}}$:
  \[
  p(z_{d,i}=k\mid\cdot)
  \propto
  \left(
  n_{d,k}^{-i}
  +
  \alpha_d\widehat\beta_k^{-i}
  \right)
  p_k^{-(d,i)}(e_{d,i}),
  \]
If a new group is spawned, we next sample the component this new group maps to, setting $z_{d,i}$:
\begin{equation}
\label{eq:new-component-link}
p(z_{d,i}=k\mid \ell_{d,i}=\mathrm{new},\cdot)\propto
\begin{cases}
M_k^{-i} \, p_k^{-(d,i)}(e_{d,i}), & k=1,\ldots,K \, (\mathrm{existing})\\[4pt]
\gamma \,p_0(e_{d,i}), & k=\mathrm{new}
\end{cases}
\end{equation}

The mixing weights used in retrieval are estimates derived from the group-to-component mapping counts, which are posterior means under the partition structure:
\[
  \widehat\beta_k
  =
  \frac{M_k}
  {\gamma + \sum_{j=1}^{K}M_{j}},
  \qquad
  \widehat\beta_{\mathrm{new}}
  =
  \frac{\gamma}
  {\gamma + \sum_{j=1}^{K}M_{j}}
  \]
  \[
  \widehat\pi_{d,k}
  =
  \frac{
  n_{d,k}+\alpha_d\widehat\beta_k
  }{
  \alpha_d + \sum_{j=1}^{K}n_{d,j}
  },
  \qquad
  \widehat\pi_{d,\mathrm{new}}
  =
  \frac{
  \alpha_d\widehat\beta_{\mathrm{new}}
  }{
  \alpha_d + \sum_{j=1}^{K}n_{d,j}
  }
  \]

\section{Implementation and Reproducibility}\label{app:impl}

\subsection{Behavior Mining Procedure and Corpus}\label{app:behavior-corpus}

We construct a separate behavior corpus for each language model, Qwen3-32B and Granite-4.2-30B, using the same model for behavior mining and evaluation (i.e. the behaviors are on-policy for the respective model). We randomly sampled 20,000 mathematics prompts from NuminaMath-CoT \citep{li2024numinamath} and 20,000 physics prompts from Camel Physics~\citep{li2023camel}.

Mining follows a three-stage metacognitive procedure, inspired by \cite{didolkar2025metacognitive}. First, the model solves each problem by generating a chain-of-thought solution. It then reflects on the solution, assessing its correctness and
identifying missing or reusable reasoning behaviors. Finally, it extracts these behaviors as structured \{name: description\}
records. Extraction is performed regardless of whether the generated answer is correct or not. The extraction prompt instructs the model to replace
problem-specific numerical values with symbolic placeholders, so that the descriptions are generalizable and transferable strategies are expressed.

After extraction is performed over the full prompt set, we deduplicate behavior records based on exact match of their labels (the \texttt{name} field). The resulting Qwen corpus contains
57,955 mathematics behaviors and 75,324 physics behaviors, while the
Granite corpus contains 26,305 mathematics behaviors and 15,839 physics
behaviors; the difference reflects this deduplication step. In warm-started comparisons for a given language model, all memory-based methods use the same model-specific corpus, for standardized comparison.

We embed behavior descriptions using
\texttt{granite-embedding-english-r2} \citep{awasthy2025graniteembeddingr2models} and fit a separate PCA transform
on each model's mined corpus, reducing the embeddings to $D=128$
dimensions. Each of these PCA transforms is held fixed during fitting of the mixture as well as test-time adaptation.

\subsection{Mixture Hyperparameters}
\label{app:impl-hparams}

We fit a separate HDP-GMM for each model's corpus using the configuration detailed in
Table~\ref{tab:hparams}, using the Gibbs sampling procedure described in Section \ref{sec:gibbs} and Appendix \ref{app:gibbs}. The resulting fitted mixtures contain 64 components for Qwen and 62 for Granite, without setting any fixed component numbers a priori. The prior scale
$\Psi_0$ is set relative to the empirical covariance of the projected
behavior embeddings. The main results use explicit domain conditioning (described in Section \ref{sec:method-ttt}), where the projected query is assigned to its nearest domain centroid, and retrieval combines the mixture weights for that domain with the shared component predictive densities. In both the static and the online settings (Tables \ref{tab:baselines} and \ref{tab:main-result}), we use the top-$k'=5$ components based on normalized weights.

\begin{table}[h]
\caption{Hyperparameters for HDP-GMM fitting and retrieval, as well as the refit window. The refit settings only apply to the settings where periodic
refitting is used.}
\label{tab:hparams}
\centering
\small
\begin{tabular}{@{}p{0.45\linewidth}p{0.50\linewidth}@{}}
\toprule
Parameter & Value \\
\midrule
Global concentration $\gamma$ & 5.0 \\
Domain concentration $\alpha_d$ & 1.0 \\
NIW scale $\Psi_0$ & $1.2\times$ empirical covariance in PCA space \\
NIW degrees of freedom $\nu_0$ & $D+2=130$ \\
NIW mean strength $\kappa_0$ & 0.01 \\
Projection dimension $D$ & 128 \\
Offline Gibbs steps / burn-in & 100 / 50 \\
Novelty threshold $\tau_{\mathrm{new}}$ & 0.5 \\
Symmetric KL merge threshold $\epsilon$ & 0.1 \\
Refit interval $g$ & 10 (AIME'25, PhysReason-mini);\newline
50 (Omni-MATH, PhysReason-full) \\
Retrieval (static and online) & Top-$k'=5$ \\
Domain conditioning & Nearest-centroid classifier (explicit) \\
\bottomrule
\end{tabular}
\end{table}

\subsection{Learning the Query Projector \texorpdfstring{($f_{\theta}$)}{(f-theta)}}
\label{app:impl-ftheta}

The learned query projector $f_{\theta}$ maps the problem's embedding into the (PCA-transformed) space in which the behavior mixture was fitted. We use a multilayer perceptron with two hidden layers (\texttt{width}=[512, 256]) and a 128-dimensional output. The training data of (query, behavior) pairs comes from the mining procedure in Appendix \ref{app:behavior-corpus}; for a pair $(x,b)$, let
$q(x)=\cE(x)$ be the raw query embedding and let
$u_b=\operatorname{PCA}(\cE(b))$ be the projected behavior embedding.

Because downstream retrieval evaluates a Student-t posterior predictive density under the fitted mixture, which depends on the scale of $f_{\theta}(q(x))$ relative to each component, we train $f_{\theta}$ with a mean squared error objective over the training pairs $\mathcal{T}$ to train its output to match the coordinates and scale of the behavior embeddings:
\[
\mathcal{L}(\theta)
=
\frac{1}{|\mathcal{T}|}
\sum_{(x,b)\in\mathcal{T}}
\left\lVert f_{\theta}(q(x))-u_b\right\rVert_2^2.
\]

We learn such a projector for the respective language models, so one each for Qwen and Granite, ensuring that projected queries and the stored behavior embeddings are expressed in a shared space. Our training is performed with AdamW for 10 epochs with a batch size of 256, learning rate of $1\times10^{-4}$, and
weight decay of $10^{-4}$. The resulting projector $f_{\theta}$, which is just $0.56M$ parameters, is held fixed during evaluation, with minimal overhead.

\subsection{Cold-Start Setting and Evaluation}

In the cold-start setting, we no longer provide the language model with access to offline-mined behaviors, and we do not have $f_{\theta}$. This means that we directly operate in the raw embedding dimension (768), but since Gibbs steps occur over a smaller corpus (fewer behaviors are mined over the test set compared to the offline mining setting), this remains computationally feasible. Since there is no empirical covariance to work from or PCA performed, we set $\Psi_0 = I_{768}$, and $\nu_0=D+2 = 770$. Domain-conditioning does not arise since we treat the samples in a test set as belonging to a single domain, so this is analogous to the non-hierarchical construction.

For evaluation (in both the cold-start setting as well as with the offline-mined memory), all models are given a thinking budget of $64$k tokens, and generations are sampled with temperature $0.7$. For both settings, memory is reset following each evaluation run -- for the cold-start setting, this ensures every benchmark is run without memory, while for the offline-mined memory setting, we revert back to the "pre-trained" mixture fit on the offline behavior corpus.

\section{Complete Baseline Results}\label{app:static-results}

We include the full static-retrieval results against the baselines in Table~\ref{tab:baselines}, complementing the results with Qwen3-32B in Figure~\ref{fig:static-retrieval}. We find that with Granite-4.2-30B, Bayesian Cheatsheet outperforms the baselines again across the benchmarks, with the BC-centroids variant performing comparably to BC-PP, our main method using the posterior predictive; BC-centroids outperforms BC-PP on AIME'25, while BC-PP is better on Omni-MATH, and they are tied on PhysReason.

\begin{table}[ht]
\caption{Static-retrieval results using the mined behavior corpus for Qwen3-32B and Granite-4.2-30B. OM and PR denote Omni-MATH and PhysReason, respectively. Bold and underlined values indicate the best and second-best results within each model and benchmark.}
\label{tab:baselines}
\centering
\footnotesize
\begin{tabular}{@{}lcccccc@{}}
\toprule
& \multicolumn{3}{c}{Qwen3-32B} & \multicolumn{3}{c}{Granite-4.2-30B} \\
\cmidrule(lr){2-4}\cmidrule(lr){5-7}
Method & AIME'25 & OM & PR & AIME'25 & OM & PR \\
\midrule
Base model & 71.1 & 18.1 & 16.4 & 73.3 & 24.8 & 16.4 \\
Random behavior selection & 72.2 & 18.9 & 18.6 & 73.3 & 27.5 & 16.4 \\
Direct behavior retrieval  & 72.2 & 18.6 & 23.5 & 74.4 & 28.2 & 19.5 \\
Metacognitive Reuse, $k'=5$ & \underline{75.6} & 17.8 & 21.7 & 77.8 & 28.4 & \underline{22.6} \\
\midrule
\textbf{BC-centroids (ours)} & 74.4 & \underline{20.5} & \underline{24.8} & \textbf{81.1} & \underline{29.1} & \textbf{23.9} \\
\textbf{BC-PP (ours)} & \textbf{77.8} & \textbf{22.3} & \textbf{27.0} &
\underline{80.0} & \textbf{31.6} & \textbf{23.9} \\
\bottomrule
\end{tabular}
\end{table}

\section{Ablations}

The ablations below use the Qwen3-32B language model, and consider different parts of our inference and online update procedure.

\subsection{Top-\texorpdfstring{$k'$}{k'} Retrieval}

In Table~\ref{tab:topk-ablation}, we perform a sweep over the number of retrieved components ($k'$) used in static retrieval with BC-PP (corresponding to Table \ref{tab:baselines}). Since we draw the medoid element from each component, $k'$ is also the number of behaviors added to the prompt for metacognitive reasoning. We hold all other hyperparameters fixed. We adopt $k'=5$, which generally performs the best across the board. Notably, relative to the baselines in Table \ref{tab:baselines}, BC-PP with $k'=3$ would also outperform every baseline across the three benchmarks for Qwen3-32B.

\begin{table}[h]
\caption{Effect of the retrieval size $k'$ (number of components) on static BC-PP accuracy.}
\label{tab:topk-ablation}
\centering
\small
\begin{tabular}{@{}lccc@{}}
\toprule
$k'$ & AIME'25 & Omni-MATH & PhysReason \\
\midrule
1 & 72.2 & 19.5 & 24.8 \\
2 & 76.7 & 20.2 & 24.8 \\
3 & 76.7 & \textbf{22.5} & 25.2 \\
5 & \textbf{77.8} & 22.3 & \textbf{27.0 }\\
\bottomrule
\end{tabular}
\par\vspace{10pt}
\begin{minipage}{\linewidth}
\centering
\captionsetup{type=figure,font=small,skip=5pt}
\includegraphics[width=\linewidth]{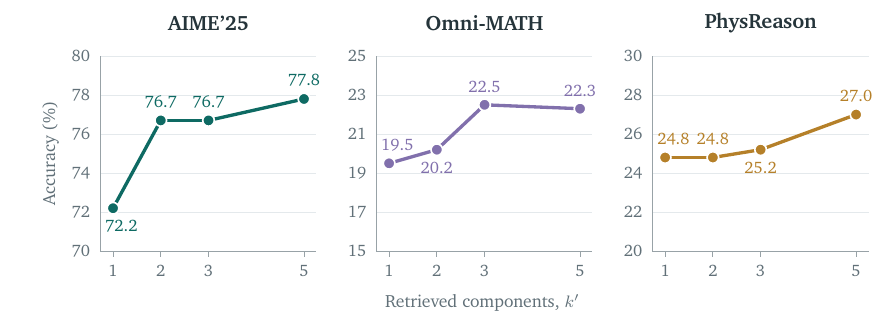}
\caption{Static BC-PP accuracy versus retrieval size $k'$ for Qwen3-32B.}
\label{fig:topk-ablation}
\end{minipage}
\end{table}

\subsection{Refit Interval Sweep}

For the online setting with refitting, we sweep the refit interval on Omni-MATH ($n = 916$), while holding $k'=5$ and the novelty threshold as well as the inference recipe (including domain conditioning) constant, and report the performance. We explore a wide range of values including $g=0$ (no refits, but new components may still be spawned) for comparison.

There are two ways that the component count can increase: 1) a new component is spawned through the novelty-driven trigger specified in Section \ref{sec:method-ttt}, and 2) a new component is formed during collapsed Gibbs resampling (refit step), so a new component effectively "splits" out of an existing one. In Table \ref{tab:refit-ablation}, we report both the total number of new components compared to the offline-fit state of the HDP-GMM ($K=64$ for Qwen3-32B) as well as the subset of \textit{novelty-induced} new components.

\begin{table}[h]
\caption{Effect of the refit interval $g$ on online Qwen3-32B accuracy on
Omni-MATH. $K$ is the final component count (initial is $64$),
``net increase'' is the net change in the number of components over the run, and ``novelty'' denotes the subset of those new components that were novelty-triggered. }
\label{tab:refit-ablation}
\centering
\small
\begin{tabular}{@{}lcccc@{}}
\toprule
$g$ & Accuracy & $K_{\mathrm{final}}$ & Net increase & Novelty-triggered \\
\midrule
0 & \underline{26.4} & 76 & 12 & 12 \\
5 & 24.8 & 94 & 30 & 5 \\
10 & 25.2 & 86 & 22 & 4 \\
25 & 25.2 & 85 & 21 & 4 \\
50 & \textbf{26.6} & 85 & 21 & 11 \\
75 & \underline{26.4} & 85 & 21 & 14 \\
100 & 25.3 & 79 & 15 & 11 \\
250 & 25.9 & 73 & 9 & 7 \\
500 & 25.9 & 75 & 11 & 10 \\
\bottomrule
\end{tabular}
\end{table}

We observe that when refits are more aggressively performed (at small values of $g$), the proportion of novelty-triggered new components to the total new components is much smaller. This suggests that most new components emerge due to the stochastic nature of the collapsed Gibbs sampling procedure, which is performed more often. As the refit interval increases, the majority of the new components are due to novelty identified over the test set pass. Based on having the highest performance and a near-even balance of novelty-triggered components and Gibbs-induced splits, we take $g=50$ for the results with refitting in Tables \ref{tab:main-result} and \ref{tab:cold-start}. We note that other values of $g$ also perform comparably: in fact, every value of $g$ tested here would at least match or outperform the strongest performing baseline for Omni-MATH in Table \ref{tab:main-result}. The $g=10$ value is used for AIME'25 and PhysReason-mini since a single pass over the test set only contains 30 and 31 samples, respectively.

\subsection{Hierarchical Structure and Domain-Conditioned Retrieval}

The inclusion of the hierarchical structure allows for domain-conditioning in retrieval through the mixture weights, permitting both cross-domain behavior sharing and potential offline domain adaptation. Does it actually make a difference in downstream performance, compared to a flat, domain-pooled DP-GMM structure? We compare the results in Table \ref{tab:hierarchical-ablation}, finding that the hierarchical structure does noticeably improve performance on all three benchmarks, suggesting that explicit domain-conditioning is indeed helpful.

\begin{table}[h]
\caption{Hierarchical (HDP-GMM) vs.\ flat, domain-pooled (DP-GMM) mixture structure for
Qwen3-32B in the online setting without refitting (all other mechanisms and hyperparameters are held fixed). The hierarchical structure clearly improves performance on all three benchmarks.}
\label{tab:hierarchical-ablation}
\centering
\footnotesize
\begin{tabular}{@{}llccc@{}}
\toprule
Mixture structure & Condition & AIME'25 & Omni-MATH & PhysReason \\
\midrule
HDP-GMM (hierarchical)
 & Online, no refit    & \textbf{81.1} & \textbf{26.4} & \textbf{30.1} \\
\midrule
DP-GMM (pooled)
 & Online, no refit  & 76.7 & 24.0 & 27.3 \\
\bottomrule
\end{tabular}
\end{table}

\subsection{Component Creation and Merging}\label{app:comp-creation-merge}

We also consider the protocols for new component creation, as well as consolidation, in the online test-time training setting. As described in Section \ref{sec:method-ttt} (Eq. \ref{eq:novelty}), if $\rho_{\mathrm{new}}(e_t) > \tau_{\mathrm{new}}$, we instantiate a new component -- this is analogous to deterministic cluster creation used in DP-means \citep{kulis2012dpmeans}. We alternatively consider a sampling scheme where we draw a $\mathrm{Bernoulli}(\rho_{\mathrm{new}}(e_t))$ variable at each step and instantiate a new component when it takes the value $1$. This procedure is faithful to the component-birth mechanism in the CRP, treating $\rho_{\mathrm{new}}(e_t)$ as a probability to be sampled. We also ablate the inclusion of the ability to merge a pair of components together where symmetric Gaussian KL $< \epsilon$, during the refitting stage.

We report the results in Table \ref{tab:threshold-kl-merge}, with all experiments involving periodic refitting ($g=50$). We find that the current configuration with thresholding and KL-based merging together performs slightly better, but the combination of Bernoulli sampling without the heuristic merge also is quite effective, and would outperform all baselines in Table \ref{tab:main-result}.

\begin{table}[h]
\caption{Performance on Omni-MATH under four combinations of new-component creation mechanism (threshold vs. sampling) and KL-based component consolidation (with vs. without). All experiments involve periodic refitting ($g=50$).}
\label{tab:threshold-kl-merge}
\centering
\small
\begin{tabular}{@{}lc@{}}
\toprule
Configuration & Omni-MATH \\
\midrule
Threshold, KL-merge                     & \textbf{26.6} \\
Sampling, KL-merge & 26.0  \\
Threshold, no KL-merge                     & 25.9 \\
Sampling, no KL-merge               & \underline{26.2} \\
\bottomrule
\end{tabular}
\end{table}

We also report the number of final components, the net increase, and the breakdown of how those components were introduced, consistent with Table \ref{tab:refit-ablation}. These statistics (specifically for Omni-MATH evaluations) can be found in Tables \ref{tab:component-spawn-g0} and \ref{tab:component-spawn-g50}. The former analyzes the setting without refitting ($g=0$), to isolate the increase in components, as they cannot be created through running a step of collapsed Gibbs sampling. For ``threshold'', all new components are novelty-triggered, while for ``sampling'', they are introduced through the Bernoulli draw. The latter includes refitting ($g=50$), reflecting these statistics for the same experiments whose benchmark performances are in Table \ref{tab:threshold-kl-merge}.

\begin{table}[h]
\caption{Component count and change on Omni-MATH without periodic refitting ($g=0$), comparing threshold vs. sampling-based component creation mechanisms. All component births are dependent on the component creation mechanism. }
\label{tab:component-spawn-g0}
\centering
\small
\begin{tabular}{@{}lcc@{}}
\toprule
Protocol  & $K_{\mathrm{final}}$ & Net increase  \\
\midrule
Threshold & 76 & +12 (novelty-triggered) \\
Sampling  & 138 & +74 (Bernoulli-triggered) \\
\bottomrule
\end{tabular}
\end{table}

\begin{table}[h]
\caption{Component counts and changes on Omni-MATH with periodic refitting ($g=50$), comparing threshold vs. sampling-based component creation mechanisms. }
\label{tab:component-spawn-g50}
\centering
\small
\begin{tabular}{@{}lcccc@{}}
\toprule
Configuration & $K_{\mathrm{final}}$ & Net increase & Online-spawned & Refit-triggered (Net) \\
\midrule
Threshold, KL-merge                    & 85 & +21 & 11 & 10 \\
Sampling, KL-merge &  129 & +65 & 69 & $-4$ \\
Threshold, no KL-merge                   & 82 & +18 & 2  & 16 \\
Sampling, no KL-merge              & 130 & +66 & 68 & $-2$ \\
\bottomrule
\end{tabular}
\end{table}

We find that the Bernoulli-based sampling procedure spawns many more components than the threshold-based method, both in isolation ($g=0$) and with refitting.  Notably, while the final number of components decreases slightly without KL-merge in the thresholding method, there are fewer (2, compared to 11) novelty-triggered component births. Thus, merging results in a higher novelty score for subsequent points by way of condensing coverage, which leads to new components being created through the $\tau_{\mathrm{new}}$ threshold. Even without the KL-based component merging, the Gibbs refit serves to collapse some of the components; with more components being created in the sampling variant, refit actually contributes a net decrease in components created.

\subsection{Novelty Threshold}

Our main experiments use $\tau_{\mathrm{new}} = 0.5$; we ablate the value of $\tau_{\mathrm{new}}$ in Table \ref{tab:tau-new-sweep}. We find that for small values of $\tau_{\mathrm{new}}$, the novelty score consistently exceeds the threshold, aggressively creating new components (212) that need to be pruned through the periodic Gibbs sweep. We find that this likely leads to noise in retrieval, noticeably hurting performance. As we increase it, we achieve a mix of novelty-triggered and refit-induced component births at $0.5$, while increasing it to $0.7$ results in novelty scores being too low to spawn new components, so all births are purely through the periodic refitting. Performance generally stabilizes, while peaking at $\tau_{\mathrm{new}}=0.5$; given it also balances novelty-based spawning with refit-based creation, we retain this for our main recipe.

\begin{table}[h]
\caption{Comparing the impact of the novelty threshold on Omni-MATH performance and component growth; all experiments use the main configuration (threshold-based component creation, refitting with $g=50$, and KL-based component merging during refitting). }
\label{tab:tau-new-sweep}
\centering
\small
\begin{tabular}{@{}lccccc@{}}
\toprule
$\tau_{\mathrm{new}}$ & Accuracy & $K_{\mathrm{final}}$ & Net increase & Novelty-triggered & Refit-triggered (Net)  \\
\midrule
0.1          & 24.3 & 206 & +142 & 212 & $-70$ \\
0.3          & 26.0 & 127 & +63  & 65  & $-2$  \\
0.5   & \textbf{26.6} & 85  & +21  & 11  & $+10$ \\
0.7          & 26.1 & 84  & +20  & 0   & $+20$ \\
0.9          & 26.0 & 83  & +19  & 0   & $+19$ \\
\bottomrule
\end{tabular}
\end{table}

\section{Qualitative Examples}

\subsection{Component Meta-Labels}\label{app:component-labels}

To analyze the components generated by fitting the HDP-GMM and the structure it induces (in terms of common thematic areas), we study whether generalizable meta-labels can be produced for each component. Specifically, we utilize Claude Opus 5 to produce a label given the list of behaviors assigned to a given component: the results for Qwen (Table \ref{tab:qwen-component-labels}) and Granite (Table \ref{tab:granite-component-labels}) are  below.

\begingroup
\renewcommand{\arraystretch}{1.25}
\begin{longtable}{@{}p{0.04\linewidth}p{0.12\linewidth}p{0.78\linewidth}@{}}
\caption{Short thematic labels for each of the $K=64$ HDP-GMM components fit on the Qwen3-32B mined behavior corpus.}
\label{tab:qwen-component-labels} \\
\toprule
\rowcolor{bmteal}
\textcolor{white}{\textbf{\#}} & \textcolor{white}{\textbf{Domain}} & \textcolor{white}{\textbf{Label}} \\
\midrule
\endfirsthead
\multicolumn{3}{l}{\small\itshape (continued from previous page)} \\
\toprule
\rowcolor{bmteal}
\textcolor{white}{\textbf{\#}} & \textcolor{white}{\textbf{Domain}} & \textcolor{white}{\textbf{Label}} \\
\midrule
\endhead
\bottomrule
\endfoot
 0 & \textcolor{white}{\colorbox{bmviolet}{\scriptsize\textsf{Math}}} & Unit consistency \& dimensional-analysis checks \\
\rowcolor{bmpanel} 1 & \textcolor{white}{\colorbox{bmochre}{\scriptsize\textsf{Physics}}} & Multi-method cross-checks \& literature verification \\
 2 & \textcolor{white}{\colorbox{bmviolet}{\scriptsize\textsf{Math}}} & Solution verification \& domain-restriction checks \\
\rowcolor{bmpanel} 3 & \textcolor{white}{\colorbox{bmochre}{\scriptsize\textsf{Physics}}} & Cosmology \& dark-matter physics \\
 4 & \textcolor{white}{\colorbox{bmviolet}{\scriptsize\textsf{Math}}} & Symmetry-based algebraic simplification \\
\rowcolor{bmpanel} 5 & \textcolor{white}{\colorbox{bmochre}{\scriptsize\textsf{Physics}}} & Applied materials \& optics engineering formulas \\
 6 & \textcolor{white}{\colorbox{bmochre}{\scriptsize\textsf{Physics}}} & Fluid dynamics \& flow-stability analysis \\
\rowcolor{bmpanel} 7 & \textcolor{white}{\colorbox{bmviolet}{\scriptsize\textsf{Math}}} & Algebraic equation-solving techniques \\
 8 & \textcolor{white}{\colorbox{bmochre}{\scriptsize\textsf{Physics}}} & Particle-physics conservation laws \& decay formulas \\
\rowcolor{bmpanel} 9 & \textcolor{white}{\colorbox{bmochre}{\scriptsize\textsf{Physics}}} & Empirical calibration \& uncertainty propagation \\
 10 & \textcolor{white}{\colorbox{bmviolet}{\scriptsize\textsf{Math}}} & Algebraic simplification \& symbolic manipulation \\
\rowcolor{bmpanel} 11 & \textcolor{white}{\colorbox{bmviolet}{\scriptsize\textsf{Math}}} & Multi-step problem decomposition \& dependency tracking \\
 12 & \textcolor{white}{\colorbox{bmviolet}{\scriptsize\textsf{Math}}} & Polynomial/rational-expression factoring \& proportionality \\
\rowcolor{bmpanel} 13 & \textcolor{white}{\colorbox{bmochre}{\scriptsize\textsf{Physics}}} & Dynamical-systems stability \& boundary-case verification \\
 14 & \textcolor{white}{\colorbox{bmviolet}{\scriptsize\textsf{Math}}} & Euclidean geometry theorems \& area formulas \\
\rowcolor{bmpanel} 15 & \textcolor{white}{\colorbox{bmochre}{\scriptsize\textsf{Physics}}} & Nuclear astrophysics \& stellar nucleosynthesis \\
 16 & \textcolor{white}{\colorbox{bmochre}{\scriptsize\textsf{Physics}}} & Condensed-matter magnetism \& superconductivity \\
\rowcolor{bmpanel} 17 & \textcolor{white}{\colorbox{bmviolet}{\scriptsize\textsf{Math}}} & Coordinate systems, symmetry \& 3D rotations \\
 18 & \textcolor{white}{\colorbox{bmochre}{\scriptsize\textsf{Physics}}} & Conservation laws \& flow/capacity analysis \\
\rowcolor{bmpanel} 19 & \textcolor{white}{\colorbox{bmviolet}{\scriptsize\textsf{Math}}} & Graph-theoretic counting \& distribution symmetry \\
 20 & \textcolor{white}{\colorbox{bmochre}{\scriptsize\textsf{Physics}}} & Estimation, decomposition \& bounding heuristics \\
\rowcolor{bmpanel} 21 & \textcolor{white}{\colorbox{bmviolet}{\scriptsize\textsf{Math}}} & Calculus optimization: extrema \& discriminants \\
 22 & \textcolor{white}{\colorbox{bmochre}{\scriptsize\textsf{Physics}}} & Magnetostatics \& magnetic-field formulas \\
\rowcolor{bmpanel} 23 & \textcolor{white}{\colorbox{bmochre}{\scriptsize\textsf{Physics}}} & Peer-review citation \& experimental-design practices \\
 24 & \textcolor{white}{\colorbox{bmviolet}{\scriptsize\textsf{Math}}} & Financial \& percentage word problems \\
\rowcolor{bmpanel} 25 & \textcolor{white}{\colorbox{bmviolet}{\scriptsize\textsf{Math}}} & Unit verification \& function-symmetry identities \\
 26 & \textcolor{white}{\colorbox{bmochre}{\scriptsize\textsf{Physics}}} & Cross-disciplinary tradeoff \& Bayesian reasoning \\
\rowcolor{bmpanel} 27 & \textcolor{white}{\colorbox{bmochre}{\scriptsize\textsf{Physics}}} & General problem-solving verification checks \\
 28 & \textcolor{white}{\colorbox{bmochre}{\scriptsize\textsf{Physics}}} & Astrophysics: supernovae, dark matter \& black holes \\
\rowcolor{bmpanel} 29 & \textcolor{white}{\colorbox{bmochre}{\scriptsize\textsf{Physics}}} & Fluid/thermodynamic physics \& dimensional consistency \\
 30 & \textcolor{white}{\colorbox{bmochre}{\scriptsize\textsf{Physics}}} & Topological \& condensed-matter quantum systems \\
\rowcolor{bmpanel} 31 & \textcolor{white}{\colorbox{bmochre}{\scriptsize\textsf{Physics}}} & General relativity, black holes \& cosmology \\
 32 & \textcolor{white}{\colorbox{bmochre}{\scriptsize\textsf{Physics}}} & Nuclear/particle physics \& authoritative-source verification \\
\rowcolor{bmpanel} 33 & \textcolor{white}{\colorbox{bmviolet}{\scriptsize\textsf{Math}}} & Combinatorics: subsets, Burnside's lemma \& pigeonhole \\
 34 & \textcolor{white}{\colorbox{bmviolet}{\scriptsize\textsf{Math}}} & Coordinate geometry: lines, conics \& perpendicularity \\
\rowcolor{bmpanel} 35 & \textcolor{white}{\colorbox{bmviolet}{\scriptsize\textsf{Math}}} & Percentage/ratio word problems \\
 36 & \textcolor{white}{\colorbox{bmochre}{\scriptsize\textsf{Physics}}} & Quantum information \& computation protocols \\
\rowcolor{bmpanel} 37 & \textcolor{white}{\colorbox{bmochre}{\scriptsize\textsf{Physics}}} & Physics heuristics \& fluid/thermodynamic formulas \\
 38 & \textcolor{white}{\colorbox{bmochre}{\scriptsize\textsf{Physics}}} & Materials science: nanoparticles \& phonon scattering \\
\rowcolor{bmpanel} 39 & \textcolor{white}{\colorbox{bmviolet}{\scriptsize\textsf{Math}}} & Combinatorial game theory \& optimal strategy \\
 40 & \textcolor{white}{\colorbox{bmviolet}{\scriptsize\textsf{Math}}} & Geometry: circles, loci \& optical formulas \\
\rowcolor{bmpanel} 41 & \textcolor{white}{\colorbox{bmochre}{\scriptsize\textsf{Physics}}} & Problem-solving meta-strategies \& memorized constants \\
 42 & \textcolor{white}{\colorbox{bmviolet}{\scriptsize\textsf{Math}}} & Arithmetic: place value, fractions \& percentages \\
\rowcolor{bmpanel} 43 & \textcolor{white}{\colorbox{bmochre}{\scriptsize\textsf{Physics}}} & Cross-domain dimensional analysis (physics/chemistry) \\
 44 & \textcolor{white}{\colorbox{bmochre}{\scriptsize\textsf{Physics}}} & Special relativity: time dilation \& Lorentz transforms \\
\rowcolor{bmpanel} 45 & \textcolor{white}{\colorbox{bmochre}{\scriptsize\textsf{Physics}}} & Biophysics \& fluid/cell mechanics \\
 46 & \textcolor{white}{\colorbox{bmochre}{\scriptsize\textsf{Physics}}} & Condensed matter: band structure \& Hall effect \\
\rowcolor{bmpanel} 47 & \textcolor{white}{\colorbox{bmochre}{\scriptsize\textsf{Physics}}} & Thermodynamics: phase changes \& adiabatic processes \\
 48 & \textcolor{white}{\colorbox{bmviolet}{\scriptsize\textsf{Math}}} & Logic \& proof techniques \\
\rowcolor{bmpanel} 49 & \textcolor{white}{\colorbox{bmviolet}{\scriptsize\textsf{Math}}} & Geometry/trigonometry fundamentals (circles, hyperbolas) \\
 50 & \textcolor{white}{\colorbox{bmochre}{\scriptsize\textsf{Physics}}} & Optics: lenses, mirrors \& convection \\
\rowcolor{bmpanel} 51 & \textcolor{white}{\colorbox{bmochre}{\scriptsize\textsf{Physics}}} & Atomic physics: quantum numbers \& spectra \\
 52 & \textcolor{white}{\colorbox{bmochre}{\scriptsize\textsf{Physics}}} & Biophysics/biochemistry: membranes \& virial coefficients \\
\rowcolor{bmpanel} 53 & \textcolor{white}{\colorbox{bmviolet}{\scriptsize\textsf{Math}}} & Probability: independence \& conditional rules \\
 54 & \textcolor{white}{\colorbox{bmviolet}{\scriptsize\textsf{Math}}} & Number theory: modular arithmetic \& divisibility \\
\rowcolor{bmpanel} 55 & \textcolor{white}{\colorbox{bmviolet}{\scriptsize\textsf{Math}}} & Calculus: derivatives, optimization \& monotonicity \\
 56 & \textcolor{white}{\colorbox{bmochre}{\scriptsize\textsf{Physics}}} & Symbolic problem-solving with placeholder variables \\
\rowcolor{bmpanel} 57 & \textcolor{white}{\colorbox{bmochre}{\scriptsize\textsf{Physics}}} & Solid-state/thermal physics \& perturbation theory \\
 58 & \textcolor{white}{\colorbox{bmviolet}{\scriptsize\textsf{Math}}} & Number theory: primality, divisors \& CRT \\
\rowcolor{bmpanel} 59 & \textcolor{white}{\colorbox{bmochre}{\scriptsize\textsf{Physics}}} & Astrophysics: gravitational waves \& black holes \\
 60 & \textcolor{white}{\colorbox{bmviolet}{\scriptsize\textsf{Math}}} & Algebra: Vieta's formulas \& symmetric polynomials \\
\rowcolor{bmpanel} 61 & \textcolor{white}{\colorbox{bmviolet}{\scriptsize\textsf{Math}}} & Statistics: central tendency \& distributions \\
 62 & \textcolor{white}{\colorbox{bmochre}{\scriptsize\textsf{Physics}}} & Theoretical physics: QFT \& string theory \\
\rowcolor{bmpanel} 63 & \textcolor{white}{\colorbox{bmviolet}{\scriptsize\textsf{Math}}} & Integration techniques \& trigonometric identities \\
\end{longtable}
\endgroup

\vspace{2cm}

\begingroup
\renewcommand{\arraystretch}{1.25}
\begin{longtable}{@{}p{0.04\linewidth}p{0.12\linewidth}p{0.78\linewidth}@{}}
\caption{Short thematic labels for each of the $K=62$ components in the HDP-GMM memory module fit from the Granite-4.2-30B-mined behavior corpus.}
\label{tab:granite-component-labels} \\
\toprule
\rowcolor{bmteal}
\textcolor{white}{\textbf{\#}} & \textcolor{white}{\textbf{Domain}} & \textcolor{white}{\textbf{Label}} \\
\midrule
\endfirsthead
\multicolumn{3}{l}{\small\itshape (continued from previous page)} \\
\toprule
\rowcolor{bmteal}
\textcolor{white}{\textbf{\#}} & \textcolor{white}{\textbf{Domain}} & \textcolor{white}{\textbf{Label}} \\
\midrule
\endhead
\bottomrule
\endfoot
 0 & \textcolor{white}{\colorbox{bmochre}{\scriptsize\textsf{Physics}}} & Chaos theory \& dynamical-systems criteria \\
\rowcolor{bmpanel} 1 & \textcolor{white}{\colorbox{bmviolet}{\scriptsize\textsf{Math}}} & Trigonometric identities in optics \& triangle geometry \\
 2 & \textcolor{white}{\colorbox{bmviolet}{\scriptsize\textsf{Math}}} & Combinatorics: permutations \& circular arrangements \\
\rowcolor{bmpanel} 3 & \textcolor{white}{\colorbox{bmochre}{\scriptsize\textsf{Physics}}} & Condensed-matter defect \& superconductor classification \\
 4 & \textcolor{white}{\colorbox{bmochre}{\scriptsize\textsf{Physics}}} & Energy-efficiency metric definition protocols \\
\rowcolor{bmpanel} 5 & \textcolor{white}{\colorbox{bmviolet}{\scriptsize\textsf{Math}}} & Combinatorial game theory \& parity arguments \\
 6 & \textcolor{white}{\colorbox{bmochre}{\scriptsize\textsf{Physics}}} & Cryptographic protocol layering \& classification \\
\rowcolor{bmpanel} 7 & \textcolor{white}{\colorbox{bmochre}{\scriptsize\textsf{Physics}}} & Radiation dosimetry \& hazard classification \\
 8 & \textcolor{white}{\colorbox{bmochre}{\scriptsize\textsf{Physics}}} & Quantum field theory formalism \\
\rowcolor{bmpanel} 9 & \textcolor{white}{\colorbox{bmviolet}{\scriptsize\textsf{Math}}} & Solid geometry: coordinates, area \& volume \\
 10 & \textcolor{white}{\colorbox{bmviolet}{\scriptsize\textsf{Math}}} & Rectangle/area word problems \\
\rowcolor{bmpanel} 11 & \textcolor{white}{\colorbox{bmochre}{\scriptsize\textsf{Physics}}} & Biomechanics \& sensory-perception force models \\
 12 & \textcolor{white}{\colorbox{bmviolet}{\scriptsize\textsf{Math}}} & Circle/conic geometry: tangents \& distances \\
\rowcolor{bmpanel} 13 & \textcolor{white}{\colorbox{bmochre}{\scriptsize\textsf{Physics}}} & Atomic transitions \& qubit-state definitions \\
 14 & \textcolor{white}{\colorbox{bmochre}{\scriptsize\textsf{Physics}}} & Nanostructure electronic transport \& quantum Hall effect \\
\rowcolor{bmpanel} 15 & \textcolor{white}{\colorbox{bmviolet}{\scriptsize\textsf{Math}}} & Solution-verification protocols \& boundary conditions \\
 16 & \textcolor{white}{\colorbox{bmochre}{\scriptsize\textsf{Physics}}} & Astrophysics: gravitational-wave \& stellar-remnant scaling \\
\rowcolor{bmpanel} 17 & \textcolor{white}{\colorbox{bmviolet}{\scriptsize\textsf{Math}}} & Age/time word problems \& exponential decay \\
 18 & \textcolor{white}{\colorbox{bmviolet}{\scriptsize\textsf{Math}}} & Probability: equally-likely outcomes \& GF(2) spaces \\
\rowcolor{bmpanel} 19 & \textcolor{white}{\colorbox{bmochre}{\scriptsize\textsf{Physics}}} & Magnetostatics: Biot--Savart law \& synchrotron radiation \\
 20 & \textcolor{white}{\colorbox{bmviolet}{\scriptsize\textsf{Math}}} & Relative motion \& relativistic velocity addition \\
\rowcolor{bmpanel} 21 & \textcolor{white}{\colorbox{bmochre}{\scriptsize\textsf{Physics}}} & Ion-channel biophysics \& membrane potentials \\
 22 & \textcolor{white}{\colorbox{bmviolet}{\scriptsize\textsf{Math}}} & Arithmetic sequences \& periodic scheduling \\
\rowcolor{bmpanel} 23 & \textcolor{white}{\colorbox{bmochre}{\scriptsize\textsf{Physics}}} & Dimensional consistency in thermo/fluid dynamics \\
 24 & \textcolor{white}{\colorbox{bmviolet}{\scriptsize\textsf{Math}}} & Graph theory: degree sums \& polyhedra \\
\rowcolor{bmpanel} 25 & \textcolor{white}{\colorbox{bmviolet}{\scriptsize\textsf{Math}}} & Combinatorial geometry: line arrangements \& gears \\
 26 & \textcolor{white}{\colorbox{bmviolet}{\scriptsize\textsf{Math}}} & Function symmetry: odd/even parity arguments \\
\rowcolor{bmpanel} 27 & \textcolor{white}{\colorbox{bmviolet}{\scriptsize\textsf{Math}}} & Number-theory \& arithmetic word problems \\
 28 & \textcolor{white}{\colorbox{bmviolet}{\scriptsize\textsf{Math}}} & Descriptive statistics: mode, median \& dice probability \\
\rowcolor{bmpanel} 29 & \textcolor{white}{\colorbox{bmviolet}{\scriptsize\textsf{Math}}} & Volume/geometry word problems \\
 30 & \textcolor{white}{\colorbox{bmviolet}{\scriptsize\textsf{Math}}} & Financial word problems: sequential transactions \& tax \\
\rowcolor{bmpanel} 31 & \textcolor{white}{\colorbox{bmviolet}{\scriptsize\textsf{Math}}} & Algebra: matrices, determinants \& polynomial roots \\
 32 & \textcolor{white}{\colorbox{bmviolet}{\scriptsize\textsf{Math}}} & Algorithmic complexity \& search-strategy heuristics \\
\rowcolor{bmpanel} 33 & \textcolor{white}{\colorbox{bmviolet}{\scriptsize\textsf{Math}}} & Balance-scale, BMI \& series-convergence identities \\
 34 & \textcolor{white}{\colorbox{bmviolet}{\scriptsize\textsf{Math}}} & Statistical estimation \& election vote-counting \\
\rowcolor{bmpanel} 35 & \textcolor{white}{\colorbox{bmochre}{\scriptsize\textsf{Physics}}} & Vector sign conventions in optics \& waves \\
 36 & \textcolor{white}{\colorbox{bmviolet}{\scriptsize\textsf{Math}}} & Chemistry stoichiometry \& mixture conservation \\
\rowcolor{bmpanel} 37 & \textcolor{white}{\colorbox{bmviolet}{\scriptsize\textsf{Math}}} & Animal/leg counting \& ratio-substitution algebra \\
 38 & \textcolor{white}{\colorbox{bmviolet}{\scriptsize\textsf{Math}}} & Calculus optimization \& blackbody/standard-candle checks \\
\rowcolor{bmpanel} 39 & \textcolor{white}{\colorbox{bmochre}{\scriptsize\textsf{Physics}}} & Traffic-flow control \& capacity-constrained systems \\
 40 & \textcolor{white}{\colorbox{bmochre}{\scriptsize\textsf{Physics}}} & Condensed-matter \& particle-physics regime classification \\
\rowcolor{bmpanel} 41 & \textcolor{white}{\colorbox{bmochre}{\scriptsize\textsf{Physics}}} & Topological defects \& anyon/monopole physics \\
 42 & \textcolor{white}{\colorbox{bmochre}{\scriptsize\textsf{Physics}}} & Diffusion \& optical-fiber transport formulas \\
\rowcolor{bmpanel} 43 & \textcolor{white}{\colorbox{bmochre}{\scriptsize\textsf{Physics}}} & Space-weather \& astrobiology assessment protocols \\
 44 & \textcolor{white}{\colorbox{bmochre}{\scriptsize\textsf{Physics}}} & Nuclear physics \& technology-readiness evidence \\
\rowcolor{bmpanel} 45 & \textcolor{white}{\colorbox{bmochre}{\scriptsize\textsf{Physics}}} & Dark matter, gravity \& black-hole bounds \\
 46 & \textcolor{white}{\colorbox{bmochre}{\scriptsize\textsf{Physics}}} & Materials characterization \& attenuation formulas \\
\rowcolor{bmpanel} 47 & \textcolor{white}{\colorbox{bmochre}{\scriptsize\textsf{Physics}}} & Wave/vibration frequency analysis \\
 48 & \textcolor{white}{\colorbox{bmviolet}{\scriptsize\textsf{Math}}} & Olympiad geometry/algebra lemmas \& proof techniques \\
\rowcolor{bmpanel} 49 & \textcolor{white}{\colorbox{bmviolet}{\scriptsize\textsf{Math}}} & Grid/number-theory formulas \& metric geometry \\
 50 & \textcolor{white}{\colorbox{bmviolet}{\scriptsize\textsf{Math}}} & Geometric length \& area word problems \\
\rowcolor{bmpanel} 51 & \textcolor{white}{\colorbox{bmviolet}{\scriptsize\textsf{Math}}} & State-tracking \& cyclic-structure combinatorics \\
 52 & \textcolor{white}{\colorbox{bmochre}{\scriptsize\textsf{Physics}}} & Dark-matter \& gravitational-wave detector-signature protocols \\
\rowcolor{bmpanel} 53 & \textcolor{white}{\colorbox{bmochre}{\scriptsize\textsf{Physics}}} & Particle-physics decay \& resonance formulas \\
 54 & \textcolor{white}{\colorbox{bmochre}{\scriptsize\textsf{Physics}}} & System boundary \& scope-definition protocols \\
\rowcolor{bmpanel} 55 & \textcolor{white}{\colorbox{bmviolet}{\scriptsize\textsf{Math}}} & Logic puzzles \& multi-component reasoning \\
 56 & \textcolor{white}{\colorbox{bmochre}{\scriptsize\textsf{Physics}}} & Thermodynamics \& propulsion/fission-yield formulas \\
\rowcolor{bmpanel} 57 & \textcolor{white}{\colorbox{bmochre}{\scriptsize\textsf{Physics}}} & Domain-agnostic evaluation \& classification protocols \\
 58 & \textcolor{white}{\colorbox{bmochre}{\scriptsize\textsf{Physics}}} & Circuit \& electrostatics formulas \\
\rowcolor{bmpanel} 59 & \textcolor{white}{\colorbox{bmochre}{\scriptsize\textsf{Physics}}} & Biophysics \& semiconductor measurement protocols \\
 60 & \textcolor{white}{\colorbox{bmviolet}{\scriptsize\textsf{Math}}} & Kinematics \& geometric-probability word problems \\
\rowcolor{bmpanel} 61 & \textcolor{white}{\colorbox{bmviolet}{\scriptsize\textsf{Math}}} & Logic puzzles \& linguistic disambiguation \\
\end{longtable}
\endgroup

\subsection{Example Behaviors (Component Medoids)}\label{app:medoid-behaviors}

To illustrate the notion of a behavior obtained through the language model's self-reflection, we provide a list of a few behaviors -- specifically, the medoid elements of some of the Qwen3-32B components described in Table \ref{tab:qwen-component-labels}. We observe that the behaviors, while providing guidance for solution strategies, are still fairly general, which facilitates effective reuse at inference time.

\begingroup
\renewcommand{\arraystretch}{1.25}
\begin{longtable}{@{}p{0.04\linewidth}p{0.095\linewidth}p{0.80\linewidth}@{}}
\caption{Selected representative behaviors from a subset of the $K=64$ HDP-GMM components fit from the Qwen3-32B-mined behavior corpus (component numbers match those in Table \ref{tab:qwen-component-labels})}
\label{tab:qwen-selected-behaviors} \\
\toprule
\rowcolor{bmterra}
\textcolor{white}{\textbf{\#}} & \textcolor{white}{\textbf{Domain}} & \textcolor{white}{\textbf{Representative behavior}} \\
\midrule
\endfirsthead
\multicolumn{3}{l}{\small\itshape (continued from previous page)} \\
\toprule
\rowcolor{bmterra}
\textcolor{white}{\textbf{\#}} & \textcolor{white}{\textbf{Domain}} & \textcolor{white}{\textbf{Representative behavior}} \\
\midrule
\endhead
\bottomrule
\endfoot
 0 & \textcolor{white}{\colorbox{bmviolet}{\scriptsize\textsf{Math}}} & Always check and convert units to ensure consistency before performing calculations, using known conversion factors if necessary. \\
\rowcolor{bmterrapanel} 4 & \textcolor{white}{\colorbox{bmviolet}{\scriptsize\textsf{Math}}} & When solving equations, check for symmetry in variables (e.g., if swapping x and y yields the same equation) to simplify or identify solutions. \\
 8 & \textcolor{white}{\colorbox{bmochre}{\scriptsize\textsf{Physics}}} & Use symmetry to simplify field contributions: opposite sides of a loop may cancel if their contributions are equal and opposite, but verify directions to avoid assuming cancellation. \\
\rowcolor{bmterrapanel} 14 & \textcolor{white}{\colorbox{bmviolet}{\scriptsize\textsf{Math}}} & When solving for area, cross-check the result with perimeter calculations (if relevant) to ensure logical consistency. \\
 18 & \textcolor{white}{\colorbox{bmochre}{\scriptsize\textsf{Physics}}} & Use multiple independent sources or methods to confirm the value of critical parameters (e.g., particle masses) before finalizing a solution. \\
\rowcolor{bmterrapanel} 30 & \textcolor{white}{\colorbox{bmochre}{\scriptsize\textsf{Physics}}} & When analyzing perturbations, always check if they break a symmetry that protects the system's properties. \\
 37 & \textcolor{white}{\colorbox{bmochre}{\scriptsize\textsf{Physics}}} & When solving for critical points (e.g., minima/maxima) in equations, use known physical constraints (e.g., SOFAR channel depth) to simplify calculations and avoid unphysical solutions. \\
\rowcolor{bmterrapanel} 39 & \textcolor{white}{\colorbox{bmviolet}{\scriptsize\textsf{Math}}} & In strategic games, determine if a player has a forced win by identifying whether the opponent can always block or avoid the winning condition regardless of moves. \\
 41 & \textcolor{white}{\colorbox{bmochre}{\scriptsize\textsf{Physics}}} & When solving a problem, outline the key concepts and their relevance before providing detailed analysis to ensure clarity and completeness. \\
\rowcolor{bmterrapanel} 43 & \textcolor{white}{\colorbox{bmochre}{\scriptsize\textsf{Physics}}} & Use a domain-specific checklist of experimental techniques (e.g., NMR, neutron scattering) to verify that no relevant methods are omitted when solving problems involving material property analysis. \\
 49 & \textcolor{white}{\colorbox{bmviolet}{\scriptsize\textsf{Math}}} & When solving geometric problems, always consider whether the figure is a special case (e.g., square, rhombus, rectangle), as this can simplify calculations. \\
\rowcolor{bmterrapanel} 52 & \textcolor{white}{\colorbox{bmochre}{\scriptsize\textsf{Physics}}} & Consider material properties (e.g., a, b) when solving applied problems to ensure practical feasibility. \\
\end{longtable}
\endgroup

\end{document}